\documentclass{article} 
\usepackage{iclr2025_conference,times}

\usepackage{hyperref}
\usepackage{url}
\usepackage{amsmath}
\usepackage{amssymb}
\usepackage{mathtools}
\usepackage{amsthm}
\newtheorem{definition}{Definition}
\usepackage{makecell}
\usepackage{tikz}
\usetikzlibrary{shapes,arrows, positioning}
\usepackage{graphicx} 
\usepackage{amsmath}
\usepackage{amssymb}
\usepackage{comment}
\usepackage{wrapfig}

\usepackage{microtype}
\usepackage{graphicx}
\usepackage{subfigure}
\usepackage{booktabs} 
\usepackage{longtable}
\usepackage{mdframed}
\usepackage{tcolorbox}
\tcbuselibrary{breakable}
\newtcolorbox{prompt}{standard jigsaw,opacityback=0,colframe=black,sharp corners=all,boxrule=0.3pt, breakable}
\usepackage{pgfplots}
\pgfplotsset{compat=1.16}
\usepackage{hyperref}
\usepackage{graphicx} 
\usepackage{multirow}
\usepackage{cancel}
\usepackage{xspace}

\usepackage{listings}
\usepackage{inconsolata}  

\definecolor{promptbg}{HTML}{F8F8F8}
\definecolor{promptborder}{HTML}{AAAAAA}
\definecolor{jsonkey}{HTML}{005CC5}
\definecolor{jsonstring}{HTML}{032F62}
\definecolor{jsonnumber}{HTML}{D73A49}
\definecolor{jsoncomment}{gray}{0.45}

\newcommand{\xhdr}[1]{\vspace{0em}\noindent{{\bf #1.}}}

\lstdefinelanguage{json}{
  basicstyle=\ttfamily\small,
  numbers=none,
  stepnumber=1,
  numbersep=5pt,
  showstringspaces=false,
  breaklines=true,
  frame=none,
  backgroundcolor=\color{promptbg},
  literate=
   *{0}{{{\color{jsonnumber}0}}}{1}
    {1}{{{\color{jsonnumber}1}}}{1}
    {2}{{{\color{jsonnumber}2}}}{1}
    {3}{{{\color{jsonnumber}3}}}{1}
    {4}{{{\color{jsonnumber}4}}}{1}
    {5}{{{\color{jsonnumber}5}}}{1}
    {6}{{{\color{jsonnumber}6}}}{1}
    {7}{{{\color{jsonnumber}7}}}{1}
    {8}{{{\color{jsonnumber}8}}}{1}
    {9}{{{\color{jsonnumber}9}}}{1}
    {:}{{{\color{black}:}}}{1}
    {,}{{{\color{black},}}}{1}
    {"}{{{\color{black}"}}}{1},
  morestring=[b]",
  stringstyle=\color{jsonstring},
  keywordstyle=\color{jsonkey},
  commentstyle=\color{jsoncomment}\itshape,
  keywords={true,false,null}
}

\newtcolorbox{promptbox}[1][]{
  enhanced,
  sharp corners,
  colback=promptbg,
  colframe=promptborder,
  boxrule=0.5pt,
  arc=2pt,
  boxsep=4pt,
  left=6pt,
  right=6pt,
  top=6pt,
  bottom=6pt,
  title=\textbf{Example Prompt},
  fonttitle=\bfseries\small,
  #1
}

\def\showcomments{}
\ifdefined\showcomments
    \newcommand{\amit}[1]{\textcolor{blue}{#1 --amit}}
    \newcommand{\naga}[1]{\textcolor{magenta}{#1 --naga}}
    \newcommand{\ashm}[1]{\textcolor{red}{#1 --ashmit}}
    \newcommand{\abh}[1]{\textcolor{red}{#1 --abhinav}}
    \newcommand{\sukr}[1]{\textcolor{red}{#1 --sukruta}}
    \newcommand{\navin}[1]{\textcolor{red}{#1 --navin}}
\else
   \newcommand{\amit}[1]{}
    \newcommand{\naga}[1]{}
    \newcommand{\ashm}[1]{}
    \newcommand{\abh}[1]{}
    \newcommand{\sukr}[1]{}
    \newcommand{\navin}[1]{}
\fi
\newcommand{\ldr}{\textsc{LiveDRBench}\xspace}
\newcommand{\scifacts}{\textsc{SciFacts}\xspace}
\newcommand{\datasets}{\textsc{NovelDS}\xspace}
\newcommand{\patents}{\textsc{PriorArt}\xspace}
\newcommand{\flights}{\textsc{Flights}\xspace}
\newcommand{\entities}{\textsc{Entities}\xspace}
\newcommand{\curie}{\textsc{Curie}\xspace}

\newcommand{\evalout}{Ground-Truth Claims}

\title{Characterizing Deep Research: A Benchmark and Formal Definition}
\author{Abhinav Java\thanks{Joint first authors, ordered alphabetically.}\\
Microsoft Research\\
Bengaluru, India\\
\And
Ashmit Khandelwal\footnotemark[1]\\
Microsoft Research\\
Bengaluru, India \\
\And
Sukruta Midigeshi\footnotemark[1]\\
Microsoft Research\\
Bengaluru, India \\
\AND
Aaron Halfaker\\
Microsoft\\
Redmond, USA\\
\And
Amit Deshpande\\
Microsoft Research\\
Bengaluru, India \\
\And
Navin Goyal\\
Microsoft Research\\
Bengaluru, India \\
\AND
Ankur Gupta\\
Microsoft\\
Redmond, USA\\
\And
Nagarajan Natarajan\thanks{Joint advising. Correspondence to: \texttt{nagarajn@microsoft.com}, \texttt{amshar@microsoft.com}}\\
Microsoft Research\\
Bengaluru, India \\
\And
Amit Sharma\footnotemark[2]\\
Microsoft Research\\
Bengaluru, India\\
}
\iclrfinalcopy
\begin{document}

\maketitle
\begin{abstract}
    Information tasks such as writing surveys or analytical reports require complex search and reasoning, and have recently been grouped under the umbrella of \textit{deep research} --- a term also adopted by recent models targeting these capabilities. Despite growing interest, the scope of the deep research task remains underdefined and its distinction from other reasoning-intensive problems is poorly understood. In this paper, we propose a formal characterization of the deep research (DR) task and introduce a benchmark to evaluate the performance of DR systems. We argue that the core defining feature of deep research is not the production of lengthy report-style outputs, but rather the high fan-out over concepts required during the search process, i.e., broad and reasoning-intensive exploration.  
    To enable objective evaluation, we define DR using an intermediate output representation that encodes key claims uncovered during search—separating the reasoning challenge from surface-level report generation. 
    Based on this formulation, we propose a diverse, challenging benchmark \ldr with 100 challenging tasks over scientific topics (e.g., datasets, materials discovery, prior art search) and public interest events (e.g., flight incidents, movie awards). Across state-of-the-art DR systems, F1 score ranges between 0.02 and 0.72 for any sub-category.    OpenAI's model performs the best with an overall F1 score of 0.55. 
    Analysis of reasoning traces reveals the distribution over the number of referenced sources, branching, and backtracking events executed by current DR systems, motivating future directions for improving their search mechanisms and grounding capabilities. The benchmark is available at \url{https://github.com/microsoft/LiveDRBench}.
\end{abstract}

\section{Introduction}
Deep Research (DR) has emerged as a popular paradigm for applying the reasoning capabilities of AI systems. DR systems have been shown to be useful for writing research reports~\citep{li2025webthinker}, business reports~\citep{msresearcher-blog}, and answering needle-in-a-haystack queries~\citep{openai-blog}. These tasks involve non-trivial reasoning that may take a human expert hours to complete and the expectation is that DR systems can significantly reduce that time~\citep{futuresearch2025deepresearchbenchevaluating,openai-blog}.

\begin{figure}[th]
    \centering
    \subfigure[]{
        \includegraphics[width=0.48\textwidth]{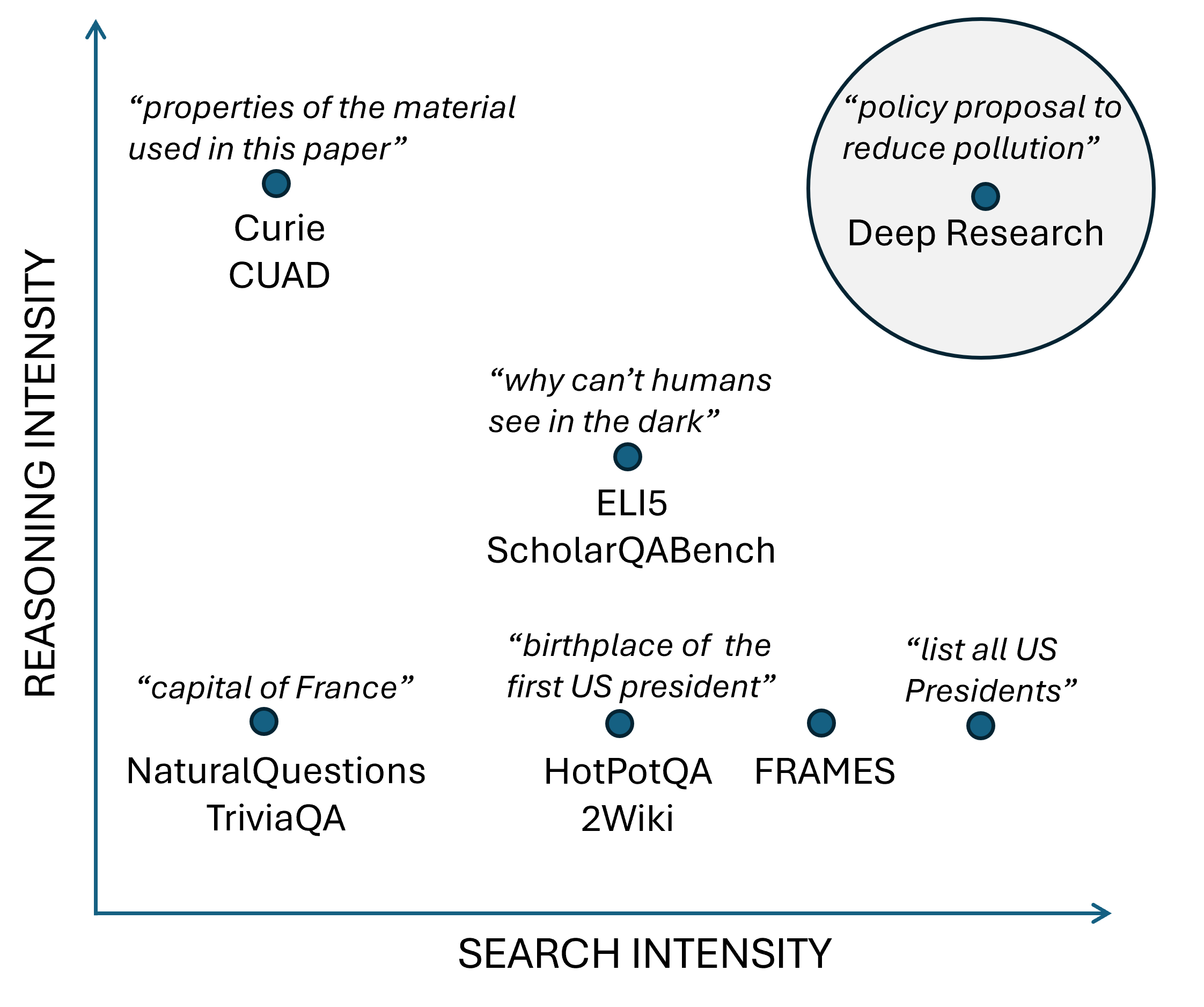}
        \label{fig:dr-def-a}
    }
    \subfigure[]{
        \includegraphics[width=0.48\textwidth]{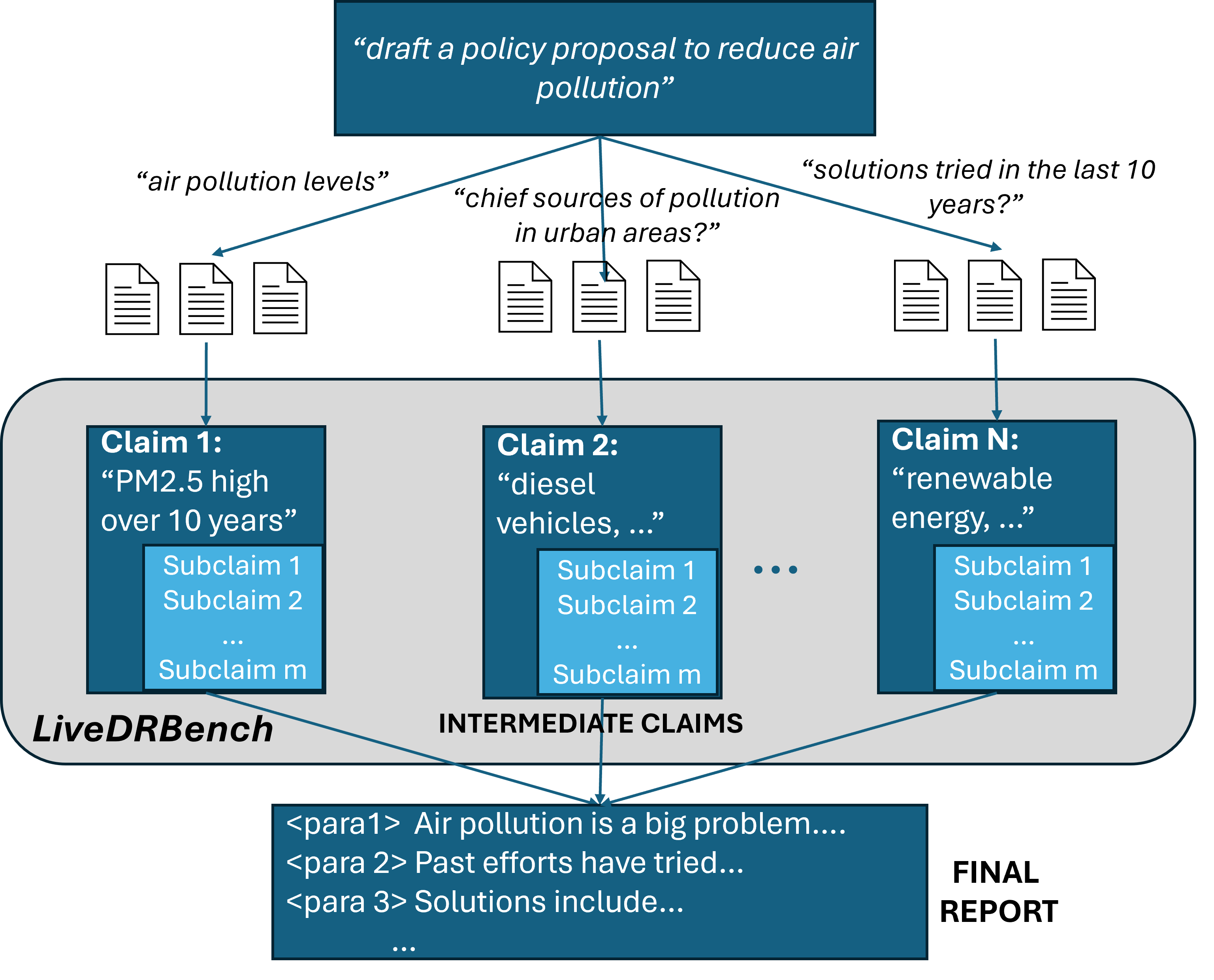}
        \label{fig:dr-def-b}
    }
    \caption{\textbf{Characterizing the deep research task.} Left (a) shows the landscape of various multi-hop reasoning tasks. Compared to existing tasks, the deep research task involves both high search  and high reasoning intensity. Right (b) shows a stylized process of generating an answer to a DR query: DR query $\rightarrow$ Claims $\rightarrow$ Long form report. \ldr focuses on the precision and completeness of the intermediate but crucial step of claim generation.}
    \label{fig:dr-def}
\end{figure}


\begin{figure}[ht]
    \centering
    \begin{minipage}[t]{0.49\linewidth}
        \centering
        \includegraphics[width=\linewidth]{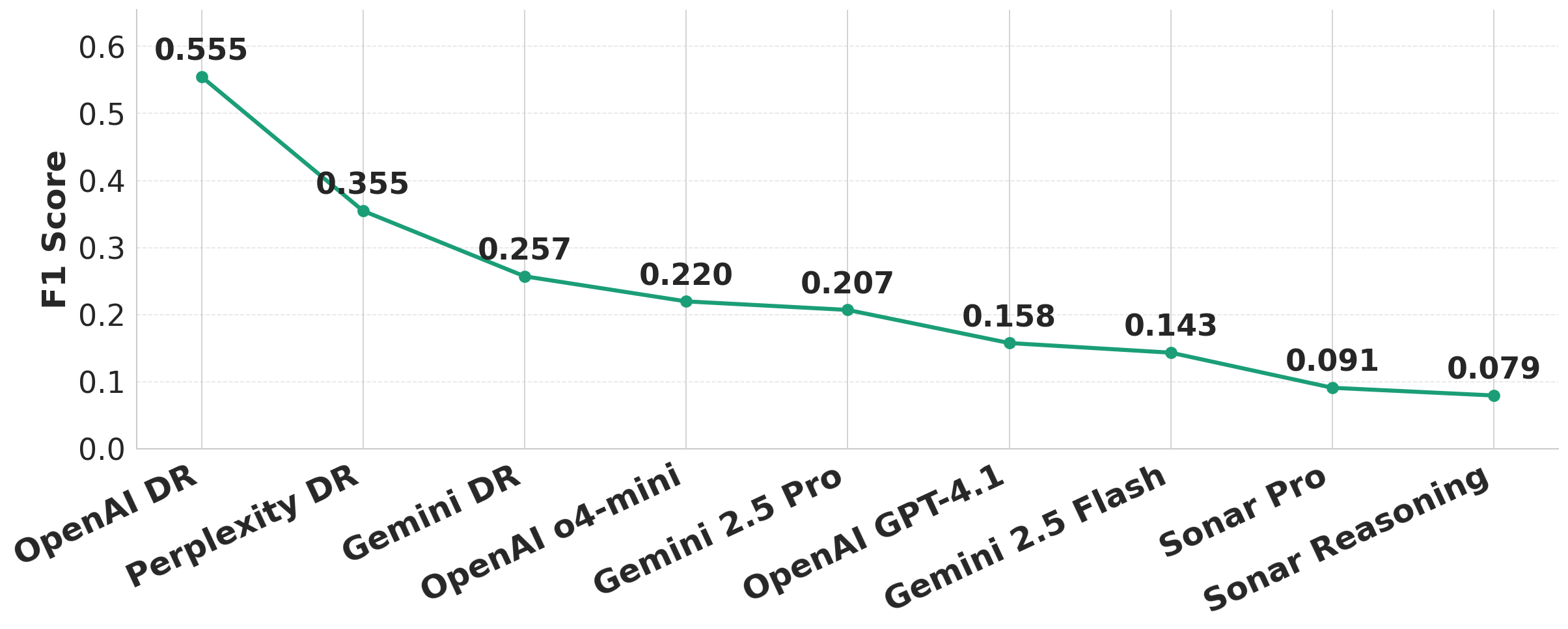}
        \caption{Average F1 score for deep research models and search-enabled LLMs on \ldr.}
        \label{fig:model-comparison}
    \end{minipage}
    \hfill
    \begin{minipage}[t]{0.49\linewidth}
        \centering
        \includegraphics[width=\linewidth]{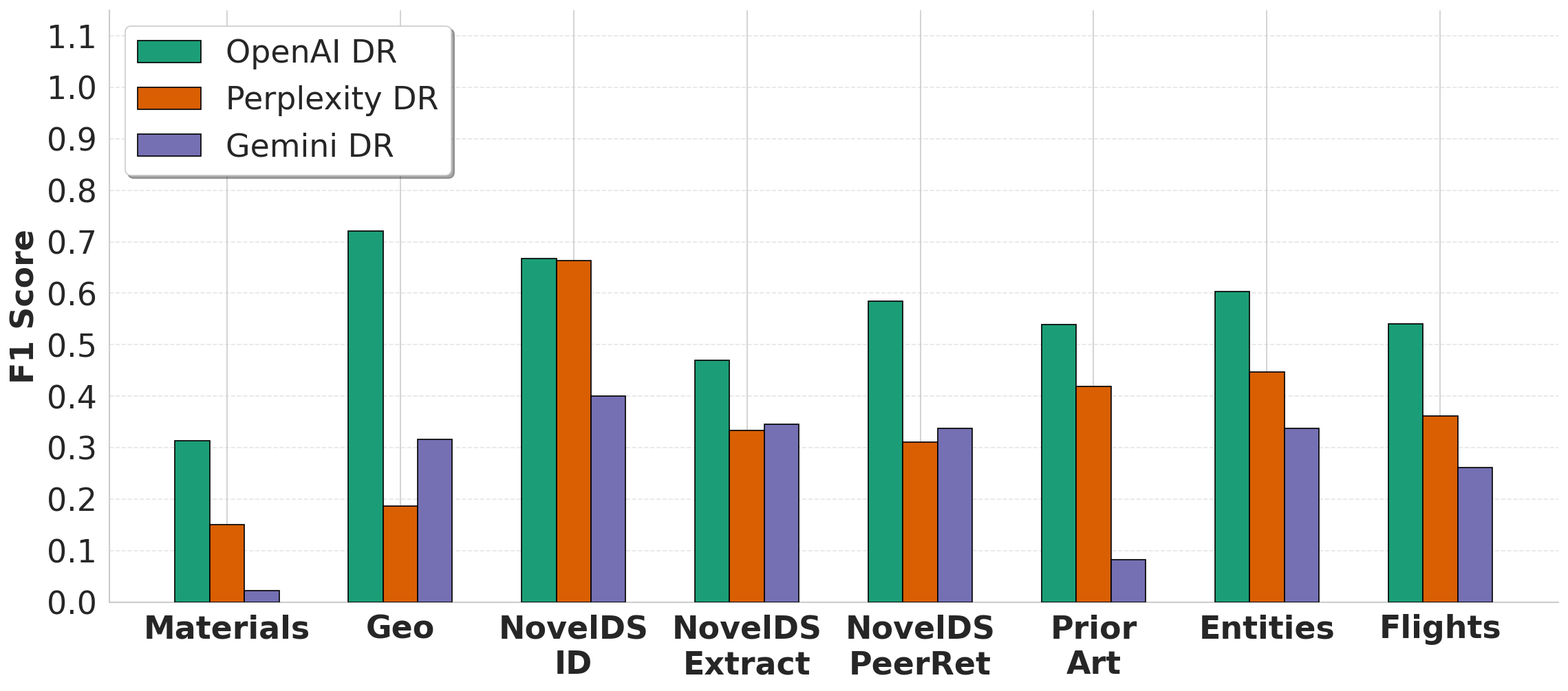}
        \caption{Average F1 scores of deep research models split by the eight categories of \ldr. OpenAI DR obtains the highest F1.}
        \label{fig:main-results}
    \end{minipage}
\end{figure}

Beyond demonstrative examples, however, no characterization of the problem exists. Typically, DR tasks are understood to be report-writing tasks that take non-trivial time for a human to complete. 
They are considered similar to reasoning-intensive search tasks, but expected to be much harder than those  studied in the literature such as multi-hop QA~\citep{hotpotqa,bright}. Interestingly, the hardness of a task also depends on the available document corpus. For example, ``provide me a list of all Oscar-winning movies that were adapted from books with women authors'' seems to be a hard, research-oriented task, but may not qualify as DR if there's a webpage providing exactly this information. Overall, the lack of a formal definition  makes it hard to evaluate DR  models and measure progress. 

In this paper, we characterize the deep research task and provide a benchmark that can evaluate progress objectively. We propose that the DR task can be split into two subtasks: 1) synthesizing \textit{claims} that collectively answer the user's query, given a document corpus; and 2) writing a report based on these claims. By a claim, we refer to any piece of information that is relevant for answering a user's query, wherein each claim may consist of sub-claims that support or provide evidence for the said information.   Rather than the common expectation of a long report-like output, we posit that the \textit{defining} element
of a DR problem is the first sub-task:  synthesis of relevant information that could answer a user's question, given a document corpus. The information synthesis problem can be conceptualized as a directed acyclic graph (DAG), where the nodes represent information such as original query, retrieved documents and synthesized information while the edges represent actions such as issuing search queries and reasoning over retrieved documents (see Figure~\ref{fig:dr-def-b}).  

Consequently, we can characterize a task by the structure of its \textit{search and reasoning} DAG from the perspective of a human expert. Given a document corpus, we call a query a deep research query if 1) the number of information units to be processed for obtaining the final answer is high (\textit{search intensity}); and 2) at least one of the tasks---finding these information units, processing them or combining them to form the final claims---requires non-trivial reasoning (\textit{reasoning intensity}). An information unit roughly corresponds to an atomic piece of information, such as a paragraph or a chunk in a retrieval corpus. 
While quantifying search or reasoning intensity is subjective, we posit that DR corresponds to any query that takes more than 10 minutes for an ideal human expert. 
Assuming the entire web as the retrieval corpus, example problems include \textit{``all oscar-winning movies adapted from books by women authors with runtime 80-95 minutes''}  and \textit{``find a material having the following properties''}. Examples of non-DR queries include ``properties of the material used in this paper'' (low search intensity) and ``list of papers accepted at NeurIPS 2024'' (low reasoning intensity).  

The above definition allows us to distinguish DR from commonly studied reasoning-intensive information tasks~\citep{frames,hotpotqa,2wiki,bright} in terms of the two axes:  search and reasoning intensity (see Figure~\ref{fig:dr-def-a}). 
Typical multi-hop QA  benchmarks such as HotpotQA~\citep{hotpotqa} contain queries with low search and low reasoning intensity. Tasks such as scientific or legal QA given a document such as \curie~\citep{curie} and CUAD~\citep{hendrycks2021cuad} respectively are reasoning-intensive but do not require a non-trivial search process. Other tasks such as explain-like-im-5 (ELI5)~\citep{eli5} involve non-trivial reasoning but the search intensity may be limited since the question asks to explain a specific concept. In contrast, most DR tasks are the ones that are both search- and reasoning-intensive. 
For example, reasoning  benchmarks such as GAIA~\citep{gaia} and  humanity's last exam~\citep{hle} qualify as both high search and high reasoning intensity DR tasks. Besides these \textit{task-completion} benchmarks that are evaluated on a single final answer, typical \textit{information-synthesis} DR applications such as writing a policy proposal, an academic survey, or an analytical report involve complex output with many aspects. We focus on the latter in this paper.

Given the complex output, evaluating a DR model's output is difficult. In particular, assessing the quality of a detailed report is an inherently subjective task, involving evaluations of style and substance. For an objective evaluation, we evaluate a DR output only on the substantive claims that it generates. 
The second subtask of generating a coherent and  detailed report can be considered as an auxiliary long-form generation task~\citep{he2025precise}, and evaluated using the standard metrics for long-form generation. Formally, given a retrieval corpus, we define the solution to a DR problem in terms of an intermediate output representation that consists of a list of claims that answers a user's query. 
Then, a DR problem is defined as a tuple $\langle$query, list of claims$\rangle$ where an ideal solution gets all the claims and their (recursive) subclaims correct. Since users may typically evaluate both the claim and its supporting subclaims for assessing validity of the DR report, we define modified versions of Precision and Recall metrics, where the score for a claim is assessed zero if the claim is incorrect, but also if all its subclaims are incorrect.

Based on this formulation, we provide a benchmark, \ldr,  consisting of 100 challenging queries that qualify as deep research  assuming the entire web as the retrieval corpus. Unlike contemporaneous benchmarks~\citep{futuresearch2025deepresearchbenchevaluating,du2025deepresearchbenchcomprehensivebenchmark}, our benchmark construction enables easy addition of new problems to adapt to the changing web and new model releases. Benchmark queries span two domains: 1) scientific, aimed at capturing the utility of DR to scientists and information professionals; 2) world events, aimed at capturing the utility of DR for the general public. Specifically, the scientific domain contains queries to identify datasets or a material given certain properties and to identify whether a given idea is novel. The world events domain includes queries about flight crashes and entities such as books and movies. These queries span various types of DR queries, from needle-in-a-haystack to queries that require enumerating a list of entities. The benchmark is available at \url{github.com/microsoft/LiveDRBench}.


We use \ldr to evaluate state-of-the-art DR systems from OpenAI, Perplexity AI, and Google. \ldr presents challenging information-synthesis queries for these systems; the F1 score per category in these models ranges from 0.02--0.72 (see Figures~\ref{fig:model-comparison} and \ref{fig:main-results}). Overall, OpenAI's DR system obtains the highest (average) F1 of 0.55. Analysis of 
reasoning traces shows the distribution of the number of searches, branching, and backtracking events executed by these DR systems, motivating future directions for improving the search mechanisms and grounding capabilities.

\begin{table}[th]
\renewcommand{\arraystretch}{2} 
\centering
\footnotesize
\begin{tabular}{p{2.5cm}@{\hspace{0.25cm}}@{\hspace{0.25cm}}p{9cm}@{\hspace{0.25cm}}@{\hspace{0.25cm}}>{\centering\arraybackslash}p{1cm}}
\toprule
\textbf{Task} & \textbf{Description and Shortened Example} & \textbf{\# Inst.} \\
\midrule
\scifacts

Geo & Find research papers that use all datasets specified in the query.  

\textit{Find the paper(s) that use all of the datasets mentioned: American Community Survey, Flooding Data from FEMA, ...} & 19  \\
\scifacts

Materials & Find materials that match all specified measured properties.  

\textit{Find the material(s) that satisfy every one of the measured properties: Direct band gap: 3.37 eV, Exciton binding energy: 60 meV, ...} & 17  \\ \hline
\datasets 

Identification & Find a dataset matching unique characteristics and return its metadata.

\textit{Identify a dataset of long-form in-the-wild videos that are segmented into scenes with ... and user engagement signals. The dataset should ... and support analysis at the video and scene level.} & 6  \\ 
\datasets 

Identification and Extraction & Find a dataset matching unique characteristics and extract specific findings from its paper. 

\textit{I'm looking for a GPT-4 generated corpus of decisions rooted in quotidian life ... How does this corpus reveal GPT-4’s implicit generation bias across various value dimensions ...?
} & 11  \\
\datasets 

Peer Retrieval & Retrieve peer dataset papers in the same problem space based on a high-level description.

\textit{I want to compare existing 3D urban segmentation datasets ... that represent urban environments as richly annotated point clouds. How do they compare in terms of data acquisition methods such as ...?} & 3  \\ \hline
\patents & Find whether a proposed research idea has already been fully or partially explored in existing papers. 

\textit{Can you help identify if the following idea has already been done ... in other papers? 
We develop a comprehensive evaluation framework for eliciting reasoning mistakes in LLMs ... explore aspects of mistake correction in LLMs ... and explicit mistake detection in reasoning chains ... We focus on distinguishing among self-generated responses ...} & 17   \\
\hline
\entities & Find an exhaustive list of real-world entities matching detailed criteria.  

\textit{Provide a comprehensive list films that meet the following criteria:
1. ... category of animation but not anime. 2. ... Tomatometer rating of 95\% or higher. 3. .. runtime must be between 80 and 85 minutes inclusive.} & 20 \\ \hline
\flights & Find a real-world event that fits the described conditions. 

\textit{In which flight incident did a commercial airliner perform an unusually high number of go-arounds before safely landing? For each attempt, detail the time, runway, landing aids ...} &  7  \\
\midrule
\textbf{Total} & & \textbf{100} \\
\bottomrule
\end{tabular}
\caption{Tasks in \ldr, spanning eight categories.}
\label{tab:tasks-eg}
\renewcommand{\arraystretch}{1.0} 
\end{table}

\section{Background \& Related Work}
We discuss recent deep research systems and attempts at evaluating them. We also discuss earlier efforts at long-form report generation, that were a precursor to the deep research systems. Note that we focus on \textit{information synthesis} deep research tasks. While information synthesis is a core capability for DR, complex DR tasks may also include computer use~\citep{browsecomp}, writing code~\citep{wang2025geak}, or calling external tools beyond search~\citep{gaia}. 

\textbf{Deep Research task.} In late 2024, Google launched a product aimed at deep research queries~\citep{gemini-dr}. Subsequently, OpenAI launched its own deep research model in February 2025~\citep{openai-blog}, which was also integrated with Microsoft Copilot's deep research agent~\citep{msresearcher-blog}. Two other deep research systems that launched soon after include Perplexity~\citep{perplexity-dr} and Grok~\citep{grok-dr}. However, beyond qualitative examples, details on system architecture and evaluation are not available.   

In the academic literature, studies on knowledge synthesis and long-form generation can be considered as the precursor for DR systems. Asai et al.~\citep{asai2024openscholar} propose a model for the task of scientific literature search that can output long form answers. Another technique is to introduce planning~\citep{godbole2024analysis} or conversational prompts~\citep{shao2024assisting} with a given language model to improve its long-form generation capabilities. 
After the release of commercial DR systems, open source implementations of deep research have also gained momentum. Huggingface deep research~\citep{huggingface-dr-2025} was one of the earliest implementations. Subsequently, systems such as WebThinker~\citep{li2025webthinker}, Open Deep Research~\citep{dzhng-deep-research2025}, and DeerFlow~\citep{deerflow2025} have been proposed. 
These systems are commonly developed using an agentic framework~\citep{huggingface-dr-2025,dzhng-deep-research2025,deerflow2025}---involving programmatic orchestration of different LLM calls to answer the user's query---or using reinforcement learning-based finetuning~\citep{shi2025pangu,li2025webthinker,zheng2025deepresearcher}. 

Although there are extensive surveys of deep research~\citep{huang2025deepresearchagentssystematic,xu2025comprehensivesurveydeepresearch}, we find that a formal definition of the problem is lacking. We provide a definition in Section~\ref{sec:define-dr} that helps characterize the problem. 

\textbf{Evaluating Deep Research.} 
Given that typical DR output involves a detailed report, most evaluation efforts focus on the  quality of long-form text output~\citep{du2025deepresearchbenchcomprehensivebenchmark,xu2025comprehensivesurveydeepresearch,chandrahasan2025deep}. 
DeepResearch Bench \citep{du2025deepresearchbenchcomprehensivebenchmark} evaluates deep research by assessing the quality of the final report on aspects such as comprehensiveness, insight, readability, and instruction-following . However, all these metrics are computed by prompting an LLM judge such as GPT-4,  which may not be a reliable measure of output (and search) quality. ResearcherBench~\citep{xu2025researcherbench} focuses on research-relevant queries in the artificial intelligence field and uses human experts to develop custom evaluation criteria for each question. Given the manual effort, it may be difficult to extend this benchmark to other fields and types of questions.  On the other end, the FutureSearch team 
\citep{futuresearch2025deepresearchbenchevaluating} considers economically valuable DR tasks and evaluate DR systems on queries such as finding or deriving a numeric answer, gathering evidence for a query, or validating a claim. Many of their categories overlap with ours, however their dataset is not available publicly. Our key insight is problem \textit{inversion}, which allows us to convert existing datasets into deep research datasets and provides a recipe for periodically updating the benchmark, which is not provided with existing benchmarks. A recently released benchmark from OpenAI~\citep{wei2025browsecomp} also follows the problem inversion idea, but it focuses only on needle-in-a-haystack search problems.

Compared to these contemporaneous benchmarks, we provide an  objective, public benchmark that straddles problems from \textit{needle-in-a-haystack} to \textit{information-gathering} problems that involve completing a list (see examples in Table~\ref{tab:tasks-eg}). We also provide a mechanism to refresh the questions in our benchmark to guard against contamination by latest models. Our formulation of deep research evaluation through claims is motivated by work on evaluating long-form generation through a set of input claims~\citep{he2025precise} and other claim verification efforts~\citep{metropolitansky2025veritrail,dmonte2024claim}. 

Finally, a key concern is to avoid including any document that provides a direct answer to a DR query in the retrieval corpus. For the specific task of information discovery and organization over academic papers, ResearchArena~\citep{kang2024researcharena} provides a benchmark with a static retrieval corpus over publications.  DeepResearchGym \citep{coelho2025deepresearchgymfreetransparentreproducible} emphasizes reproducibility of results and develops a (static) retrieval corpus on which to execute tasks. We take a different approach: we develop our benchmark to be as close as possible to the real-world use case of DR systems. As a result, retrieving from the entire Web is allowed, and we develop specialized techniques to create queries that are still hard to answer. 

\section{Defining Deep Research}
\label{sec:define-dr}
We divide the DR task into two subtasks: 1) generating the necessary claims to write the report; 
2) writing a report based on the generated claims. 
In this paper, we focus our attention on the first subtask. Writing a report given a set of claims is a long-form text generation problem~\citep{he2025precise} that can be independently studied.

The deep research task can be considered as an \textit{extreme} version of the multi-hop RAG task. Given a document corpus, the multi-hop RAG task consists of questions such that answering each question requires retrieving and combining information from more than one document. 
There are three aspects in which the DR task is more complex than multi-hop RAG: how many pieces of information need to be processed; how easily those pieces of information are found, processed or combined; and how complex is the final output.

\textbf{Defining an information unit.} Let us first define the notion of a document more precisely. As mentioned in the Introduction, the definition of deep research (and by extension, multi-hop RAG) depends on the retrieval corpus under consideration. It is possible that a query appears multi-hop on surface, e.g., ``movie adapted from a book that won the 40th Booker Prize'', but its answer appears in a single Wikipedia page (e.g., a page about the prize-winning book also mentions the movie name upfront) and it does not qualify as a multi-hop query. In other cases, a query may involve combining information from a single document but may still be considered multi-hop. This can happen when a single document is too large (e.g., a financial report with over two hundred pages) and its different subparts (chunks) may need to be retrieved separately and combined.

Hence, we define an ``information unit'' to correspond to an atomic piece of information such as a paragraph, typically implemented as a chunk in retrieval systems~\citep{liu2024lost}.\footnote{In addition to the document size, the size of a chunk may also depend on the context window of the target language model being used for the DR task.}  Each paragraph (chunk) is considered an information unit; consequently, a single document may include multiple information units.

\subsection{Deep Research: Tasks involving both  Search and Reasoning Intensity}
The DR task extends the multi-hop RAG task in three aspects. 

\textbf{1. Number of information units that need to be processed.} Unlike multi-hop tasks that are based on 3-5 information units~\citep{hotpotqa,2wiki,frames}, many DR tasks require combining information from a large number of input sources, such as writing a survey based on tens of academic papers and content within them, or listing all items that meet a certain criteria. In terms of the search process of an ideal human expert, this corresponds roughly to the minimum number of searches issued by the expert.

\textbf{2. Complexity of finding, processing, or combining the required information units.}
In many DR tasks, the search queries to issue are not obvious from the user's question and require tens of reasoning-intensive iterations. In other DR tasks, the search queries may be easy to enumerate, but processing or combining the  retrieved information units to develop output claims is reasoning-intensive. A DR task includes at least one of the following reasoning components. 

\begin{itemize}
\item \textit{Finding information units.} For needle-in-a-haystack DR questions (such as ``find material(s) with the following properties''), identifying the right search queries based on previous queries' results is the key subproblem to be solved. 
\item \textit{Processing information units.} For other DR questions, the searches may be easy to enumerate, but processing each result may require reasoning. For example, consider the question, ``English language films since 2024 meeting the following criteria...'', where finding the listing of films is trivial but processing each film's properties needs reasoning.
\item \textit{Combining information units.} In other DR questions, the most difficult part is combining the retrieved information units to answer the user's question. For example, consider the question, ``write a survey of open-weights language models released by American companies'', where the most reasoning-intensive part is comparing information units (e.g., latency) across results of multiple searches. 
\end{itemize}

\textbf{3. Complexity of the Output. } Finally, a typical DR task requires a detailed report, involving multiple claims and their justification. We abstract out the content of the report as a nested list of claims, $\mathcal{A}$. Assuming independent top-level claims, the structure of $\mathcal{A}$ can be considered as a \textit{list of dictionaries}, where each claim corresponds to a dictionary (see Figure~\ref{fig:dr-def}b). Each top-level claim can have nested subclaims, which are the keys of the dictionary. 

Based on the above discussion, we have the following definition.



\begin{definition}\label{def:dr-full}
Given a document corpus $\mathcal{C}$, query set $\mathcal{Q}$, and an ideal human expert who does not already know the answer to any query $q \in \mathcal{Q}$,
\begin{enumerate}
\item A query $q$ is said to be a deep research query if answering it requires processing a large number of information units (search intensity) and at least one of the following subtasks---finding the required information units, processing them, or combining them---requires non-trivial reasoning (reasoning intensity). 
\item The deep research task can be formulated as a tuple $\langle$user query q,  answer list $\mathcal{A}$, corpus $\mathcal{C}$$\rangle$ where $\mathcal{A}$ consists of independent claims that make up the final answer. Each claim may recursively include sub-claims that need to be derived and presented to arrive at the claim. 
\end{enumerate}
\end{definition}

While \textit{reasoning-intensive} for a human expert is difficult to quantify, we posit that 
a problem is reasoning-intensive if the
expected time taken by the expert to solve the task using available tools such as web search is more than ten minutes. Similarly, for search intensity, a typical threshold may be 20 information units (through at least 10 search queries).  



\subsection{Evaluation: Precision and Recall over Ground-truth Claims' Structure}
\label{sec:metrics}
We present modified versions of Precision and Recall metrics that take into account the structure of the DR output as claims and subclaims. A key benefit of defining the DR output in terms of claims  is that it unites the different kinds of DR tasks. Whether asking for a list of entities, a set of datasets or a particular material, each query's answer can be broken down into claims that are necessary to provide the answer, and in turn, sub-claims that are necessary to establish each claim.

From Definition~\ref{def:dr-full}, we assume that the ground truth can be written as $\mathcal{A^*} = [A^*_1, A^*_2, ...A^*_m]$ where each $A^*_i$ is a dictionary containing subclaims for the $i$th claim. Similarly, the DR output would be a list of claims, $\mathcal{A}$. Before we define the metrics, let us set up some terminology.  

Let $s_{A_i}$ be the agreement score between any claim $A_i$  and the ground truth claim $A^*_i$. Typically, this will be a binary score $\in \{0,1\}$. For each independent claim, its contribution to the Precision metric is the product of its own agreement score and the precision  of its subclaims (which would be an average over agreement scores for all the subclaims). 
That is: 1) if the claim itself is incorrect, the final score for a claim should always be zero irrespective of its subclaims. For example, if the task is to identify a material having specific properties along with the source document, the DR system should get a score of zero if it outputs the wrong material, irrespective of the correctness of its source document; 2) Similarly, if all subclaims are incorrect, then even if the claim is correct, the output obtains a score of zero. This is to avoid high precision by simply memorizing the answer. Continuing the example, if the material is correct but the source document does not exist (or does not make sense), it should be assigned a score of zero since the answer does not follow from the DR process (and likely, such a system will falter in novel scenarios where the answers are unlikely to be memorized). 
For the recall metric, we follow the same approach.

Formally,  for each claim $A_j$, $\operatorname{subc}(A_j)$ represents its subclaims where $\operatorname{subc}(A_j)=\phi$ for an atomic claim. Then, for a DR system's output $\mathcal{A}$, we present two ways of evaluation: \textit{standard} and \textit{strict}. In the standard evaluation, we  average  the  contributions of independent claims for both precision and recall: 
\begin{equation} \label{eq:prec-recall}
 \text{Prec}(\mathcal{A})= \frac { \sum_{A_i} w_i  s(A_i) \operatorname{Prec}(A_i) } {\sum_{A_i} 1}; \text{\ \ } \text{Rec}(\mathcal{A}) = \frac { \sum_{A_i} w_i  s(A_i) \operatorname{Rec}(A_i)} {\sum_{A^*_i} 1}
\end{equation}

where $\operatorname{Prec}(A_j) = \operatorname{Rec}(A_j)= 1$ whenever $|\operatorname{subc}(A_j)|=0$. $w_i$ is an optional weight that can be assigned to each claim, e.g., to weight the most important claims higher.

The strict version of the score takes the minimum instead of the mean of the constituent claims' scores. This formulation is motivated by user trust in the final report:  a user is unlikely to trust (even a correct) claim if any of its derivatory subclaims are incorrect. 

$$ \operatorname{Prec}(\mathcal{A})= { \min_{i \in \{1,2, \dots |\mathcal{A}|\}}  w_i s(A_i) \operatorname{Prec}(A_i)    }; \text{\ \ } \text{Rec}(\mathcal{A}) =  {\min_{i \in \{1,2, \dots |\mathcal{A}^*| \}} w_i s(A_i) \operatorname{Rec}(A_i)}$$

where $\operatorname{Prec}(A_j) = \operatorname{Rec}(A_j)= 1$ when $|\operatorname{subc}(A_j)|=0$ as before. 
Effectively, such a score will be 1 only if the DR model's output exactly matches the list of dictionaries $\mathcal{A}$: no claim is incorrect and no claim is missing. Such an evaluation may be useful when asking for an enumeration of a certain class of entities, e.g., all movies that satisfy a constraint. One may consider a DR system's output a failure if it fails to retrieve exactly the number of movies that satisfy the constraint. 

Note that we assume that the DR output is a list of dictionaries $\mathcal{A}$. In case the DR system outputs a report, one can use a large language model such as GPT-4 to process the report and extract the report's key claims in this format. We use the json format in our implementation. 

\begin{figure}[t]
\includegraphics[width=\linewidth]{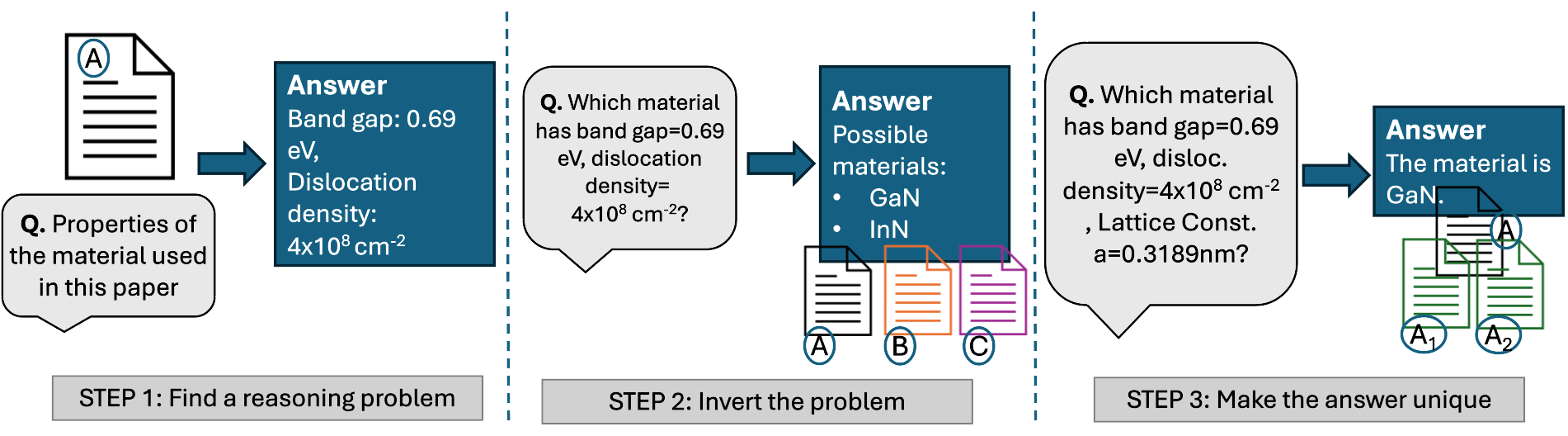}
\caption{Problem inversion process to create \ldr tasks. The first step is to find a long-context or document reasoning problem that includes a question based on the document and its ground-truth answer. In the second step, this problem is inverted to create a new question asking for an event or entity consistent with the properties mentioned in an answer. In the third step, the question is refined (e.g., more properties are added) such that it admits a unique answer. Finally, the ground-truth set of reference documents is updated in case there are additional documents ($A_1, A_2$) that provide the same answer.}
\label{fig:problem-invert}
\end{figure}

\section{Building \ldr}
In this section, we give descriptions of the tasks in our benchmark. We start with the rationale for the domains and tasks, and then describe on how we construct and evaluate the tasks. 

\subsection{Criteria for tasks}
Formulating Deep Research (DR) evaluation tasks is tricky because the models typically have access to the entire Web at inference time. We list the following desiderata for a DR evaluation benchmark:
\begin{enumerate}
\item The tasks should not be answerable using any single article on the Web. This is especially challenging, as the Web is constantly updated and the difficulty of a task depends on the corpus (i.e., if there is a new document summarizing pros and cons of a method, then the task of determining pros and cons of the method can be solved by a single retrieval and is no longer a DR task). 
\item The tasks should require accessing a wide variety of sources on the Web, i.e., to force the models to find the right grounding for their responses and not be able to  answer directly from parametric memory. 
\item Evaluation of each task should be objective and reproducible, and metrics should help benchmark the progress in the DR field.
\item The tasks should be amenable to periodic updates. This helps keep pace with fast-moving Web as well as fast-changing models.
\item The tasks should capture a wide variety of DR domains and users (e.g., scientists, researchers and analysts, common Web users).
\end{enumerate}


Striking a balance between all these constraints is not trivial. For instance, a benchmark constructed by asking domain experts to provide the questions (as in \citep{du2025deepresearchbenchcomprehensivebenchmark}) satisfies the second property, but it may not be easy to update. Instead, we propose a strategy based on \textit{inversion} of standard NLP and reasoning problems (see Figure~\ref{fig:problem-invert}). Several tasks in our benchmark are inversions of the long-context reasoning problem (e.g., \curie \citep{curie} benchmark). The standard long-context reasoning problem has the following flavor: 

[Long-context Reasoning] \textit{Given a long source document on an entity or a subject and a list of keys, identify values for the keys consistent with the document}. 

For instance, given a research article about the effects of a certain treatment, answering questions about the subjects of study, statistical significance, etc. would fall under long-context reasoning. The inverted problem becomes a DR task: 

[DR] \textit{Given keys and values that conform to a certain (unique) entity, identify the entity and locate the sources that corroborate the identity.} 

In the above example, the inverted DR task is to locate the research study article(s) given the parameters of user study, participant details, symptoms they had, and chemicals used in the treatment. Note that the DR task may yield multiple articles as the correct answer: although we started with a single research article, it is possible that the extracted dictionary of $\langle$keys, values$\rangle$ may not uniquely identify the article. In practice, we manually verify  that the DR question yields a unique answer and if not, update the extracted dictionary (or in some cases, extend our ground-truth manually by adding the additional correct answers).



Based on the aforementioned desiderata (1--5), our tasks cover the domains of science (facts and datasets), innovation (prior art search), and global culture and events (entities and flights). As discussed in the Introduction, we formulate tasks that expect claims with right grounding as output (see Figure~\ref{fig:dr-def}) and ensure that these tasks have a range of reasoning and search requirements. Using claims as intermediate outputs, we are able to use standard, objective metrics such as Precision and Recall to compare DR models. Example prompts for each benchmark category are provided in Appendix~\ref{app:examples}.

\subsection{Task Categories}

\subsubsection{\scifacts: Sourcing Scientific Facts}
We create 31 queries designed to cater to scientists and researchers. This dataset has two distinct subsets - (a) material identification, and (b) geospatial paper identification. Below we discuss both these categories in detail.

\noindent\textbf{Materials.} We create \textbf{17 questions} to emulate scenarios where a scientist might want to look for a specific \textit{compound or material} that perfectly fits a given set of criteria. Each question comprises a list of measured properties that maps perfectly to a compound; and the task of a deep research agent is to successfully identify this compound, along with the source of information (academic paper). We leverage the \emph{mpve} subset from the \curie~\citep{curie} dataset to construct a valid set of ground truth answers. The \curie~\citep{curie} dataset consists of long-form comprehension and reasoning questions. Here, a subset of questions require the model to enumerate all the measured properties of a given material from a target paper. To convert this to a deep research query we leverage the ground truth measured properties and design the task such that the job of the model is to identify the material, reference paper title, inference basis, and extract the relevant tables and passages from the reference paper. We compute final metrics based on the material name and paper title. An example of this task is provided in the Appendix~\ref{app:examples}.

\noindent\textbf{Geo.} We create \textbf{19 questions} for paper identification to mirror the scenario of a researcher attempting to search for citations that precisely utilize a given set of \textit{geospatial analysis datasets}.  The task of the deep research agent is to find papers that utilize all the datasets listed in the question. Similar to ``Material Identification" (above), we utilize \curie~\citep{curie} to construct our ground truth corpus from their \emph{geo} subset consisting of papers from the geospatial analysis field. The \textit{geo} task in \curie involves listing all datasets used in a given research paper, and we \textit{invert} this task to construct a DR question to find the paper. An example of this task is provided in the Appendix~\ref{app:examples}.



\noindent \textbf{\evalout.} In the \scifacts Geo task, the ground truth is provided as a list of paper titles which use the datasets mentioned in the original query. Similarly, in \scifacts Materials task, we provide a list of material names (\textit{claims}) and associated supporting evidence such as paper titles (\textit{subclaims}) as the ground truth and compute the metrics according to Eq.~\ref{eq:prec-recall}. Since the material identification task consists of both the main claim and the sub-claim, it is significantly harder than the geo task.

\subsubsection{\datasets: Identifying Dataset Papers}

We create 20 questions based on dataset papers, drawing from both our own domain knowledge and the dataset papers accepted at ICLR 2025. This category contains three types of tasks, (a) Dataset Identification, (b) Dataset Identification and Extraction, and (c) Dataset Peer Retrieval. Unlike the Geo category from \scifacts, here the goal is to output dataset(s) and their associated properties.

\textbf{Dataset Identification.} These questions are designed to emulate scenarios involving a researcher looking for a dataset for a specific use-case. For each dataset paper, we identify its core contributions and the aspects that distinguish it from related work, to create open-ended queries that a human might realistically pose when unaware of the dataset’s existence but seeking something with those specific characteristics. The goal is to reflect a genuine information need, rather than merely paraphrasing the abstract. The task requires the model not only to identify the appropriate dataset, but also to extract basic metadata such as its publication year, venue, and a link to the dataset. Notably, failing to account for even one of the dataset’s unique attributes often results in the retrieval of a similar yet ill-fitting alternative.
 
\textbf{Dataset Identification and Extraction.} These questions build upon the Dataset Identification category by requiring not only the identification of the correct dataset but also the extraction of specific findings from the dataset’s paper. These questions mimic those a researcher might pose when seeking results or findings for a specific use case, without being aware of the dataset’s existence. Extracting these findings may, for example, involve reasoning over results in a table, reasoning over multiple tables, or interpreting a figure in a paper.

\textbf{Peer Dataset Retrieval.} Some dataset papers include literature surveys or comparison tables highlighting peer datasets with similar goals, and slight variations in task settings. We use these sections to construct questions that reflect shared research directions across multiple datasets, without referencing any dataset by name. These questions emulate real-world exploratory scenarios in which a researcher is familiar with the broader problem space but not with specific datasets. The task requires the model to search and reason given a high-level description to find a set of candidate datasets, and then to articulate the key features that define and distinguish each one.

\textbf{\evalout.} For the \datasets Identification task, we provide the ground-truth paper title, publication year, venue, and dataset link required by the queries. Here, paper title serves as the main claim, while the year, venue, and link form the subclaims. For \datasets Identification and Extraction and \datasets Peer Retrieval, we provide a list of ground-truth findings or peer datasets required by each query. Each item in the list includes a set of main claims (e.g., “task”, “model-name”, “dataset-name”), with supporting details such as performance metrics or dataset attributes serving as subclaims. Precision, recall, and F1 scores are computed based on the overlap of the model outputs with that of the ground-truth.

\subsubsection{\patents: Deconstructing Paper Abstracts}
\label{sec:patents}
We devise 17 novel tasks motivated by the laborious problem of validating (or invalidating) claims in patent applications \cite{churnet2012patent}. Typical response to a patent application consists of a list of claims that are accepted and those that are rejected. For each rejected claim, the response also includes citations that are prior art that already demonstrate the claims in whole or in parts, thereby challenging the grounds for patenting. To mirror this real-world scenario, we create DR tasks that involve deep reasoning over synthetic research paper abstracts. 

The task is to reason over the given paper abstracts, manually written by us to mimic the paper abstracts in standard ML venues. Each abstract \textit{combines} ideas and contributions primarily from recent ML papers in ICLR 2025 and ICML 2025 conferences (including accepted and rejected papers). The goal is to understand the key ideas  in a given abstract and identify papers that were the source of those ideas, i.e., that  directly or indirectly provide evidence to support the ideas. Each task requires the model to list the claims in the abstract, references (paper titles and links), and rationales for why the references demonstrate the claims.

\textbf{Dataset creation.} We rely on OpenReview discussions to create the tasks. We start by identifying papers in ICML 2025 and ICLR 2025 venues on OpenReview where there is sufficient engagement from the reviewers and the authors during the rebuttal phase. For each of the identified papers, we start with the paraphrasing of the contributions from the meta-reviewer, reviewers, and in some cases, the authors themselves. This becomes the seed for the synthetic abstract in the task. We find that this seed is already challenging to identify, as the paraphrasing is often nuanced, highlighting aspects beyond the corresponding published paper abstract (in particular, lexical matching of the seed text with the original paper abstract wouldn't be helpful). We then go through the discussions to identify closely related papers pointed out by the reviewers. We use the discussions as well as the contributions stated in the text of the related papers to augment the synthetic abstract for the task. 

For instance, if the seed text is about a method that works under the setting of Scenario A, and the reviewers point out a closely related paper that solves the same problem in Scenario B but using a modified method, then we write the synthetic abstract to subsume both the scenarios as well as the modifications of the method as purported contributions. 

In most cases, the related work pointed out by the reviewers are papers that \textit{are not already cited by the paper at hand}. Furthermore, in some cases, we found a few other papers that also are closely related to some the claims in the abstract. In those cases, we manually added those references to the ground-truth. Overall, this forms a challenging DR problem --- the models cannot simply follow the references in the related work of one paper to get all the source papers for the task. In this way, our tasks that require deconstructing paper abstracts are more nuanced than standard DR tasks that elicit literature review or drafting related work given a topic.

\textbf{\evalout.} The ground-truth consists of top-level claims, where each claim corresponds to an idea or a contribution discussed in the abstract. Each claim is supported by the sub-claims consisting of ground-truth paper title, link, and a connection field that describes how the claim is demonstrated in the referenced paper. The DR model is expected to produce the same format in its output. 
We compute precision, recall, and F1 scores based on the overlap of the model-reported references with that of the ground-truth.


\subsubsection{\flights: Identifying Flight Incidents}

\textbf{Dataset Creation:} We construct 7 questions based on official flight investigation reports published by national aviation authorities (e.g., the NTSB, AAIB, and BEA). Each question is derived by identifying a flight incident with unique features -- such as an unusual failure mode, a rare technical error, or a non-standard response procedure. We then \textit{invert} the task: the DR question involves identifying the correct flight incident  from a high-level description, elaborating on its causes or timeline. 

These questions emulate scenarios in which members of the general public, aviation enthusiasts, or journalists seek information about notable incidents without knowing the specific flight number. As official incident reports often span hundreds of pages, the task requires processing long documents and synthesizing information across multiple sections.

\textbf{\evalout.} We provide a ground-truth list of the incidents' timelines or causes, as required by each query. Each item in the list includes a set of main claims (e.g., “timestamp”, “flight-parameter”), with supporting details such as sensor readings, flight configurations, anomaly descriptions serving as subclaims. Precision, recall, and F1 scores are computed based on the overlap of the model outputs with that of the ground-truth.

\subsubsection{\entities: Enumerating Entities with Constraints}

We devise 20 questions pertaining to global culture that require consulting a broad range of sources and cannot be accurately answered using a small set of sources on the Web. These questions focus on search intensity. They ask for an exhaustive list of entities within a given category that satisfy all specified criteria. All questions in this category have a fixed reasoning depth, ranging from 2 to 4. They are centered around real-world events such as award shows, the Olympics, book publications, and similar topics.

\textbf{Dataset creation.} We select world events that happen at a regular cadence and design questions that cannot be fully answered using any single webpage or a small set of webpages. Each question in this category requires examining a wide range of items, typically between 80 and 140, and filtering them based on the provided criteria. The final answers consist of 8 to 30 items that meet all specified requirements. 

\textbf{\evalout.} Each question requires a list of entity names (\textit{claims}) as output. We construct the ground truth using scraping scripts tailored to each world event, followed by manual verification of agent-generated answers, particularly when they are not present in the initial ground truth. After reviewing all agent responses and adding any valid missing answers, we evaluate performance using precision, recall, and F1 scores against the finalized ground truth list.

\subsection{LiveDRBench: Enabling benchmark updation with novel questions}

We created \scifacts queries using Curie~\cite{curie}, by extracting key information from scientific papers and posing queries to use this key information to retrieve the corresponding source paper. Future queries can be similarly generated by applying the same pipeline to newly published papers. 

The \datasets Peer Retrieval questions can be extended by identifying literature survey tables in newly published papers, and extracting the listed datasets along with their distinguishing features. 

For \entities, we include four categories of queries based on distinct world events — movies, book publications, international math Olympiads, and the Olympics. More queries can be generated by selecting a similar event-based category, defining a set criteria, and iteratively scraping all items within that category satisfying the defined criteria.

\subsection{Evaluation}
\label{sec:eval}
As mentioned above, we structure each task so that it expects a \textit{list of dictionaries} as the output. Each top-level dictionary corresponds to a claim and its keys correspond to sub-claims. For instance, for the material identification task in \scifacts, identifying the material name is the claim while identifying the source paper correctly is the sub-claim that supports it. For each key in a dictionary, we define the \textit{agreement} score $s\in [0,1]$ as a measure of the  accuracy of the claim represented by the key's value. 

For convenience in evaluation, we include an instruction to output a JSON object in every DR prompt.\footnote{This format instruction can be easily omitted to allow evaluation without the JSON format requirement.} This avoids the need to parse the output report. However, some DR systems such as Gemini were unable to follow the instruction and still output a report. Therefore, our evaluation pipeline includes an additional step of parsing the keys from a given report. While this can be done by prompting an LLM such as GPT-4, for the purposes of benchmark evaluation, we decided to do it manually to avoid any possibility of error. 

\noindent \textbf{Metrics.} For each benchmark category, we evaluate the correctness (\textit{precision}) of the stated claims and completeness (\textit{recall}) of those claims. We first compute the agreement score $s$ of a generated claim with its ground-truth counterpart (discussed in Section \ref{sec:metrics}) using GPT-4o (prompt included in Appendix \ref{app:evalprompts}). Then, we compute precision and recall metrics as defined in Equation~\eqref{eq:prec-recall}. We also report the F1 score combining these two metrics. 

However, there is a limitation of the above strategy. While we considered the uniqueness of the answer while creating each DR query, it is possible that multiple correct answers exist. As a result, given only the ground-truth claims, the recall metric may miss correct claims that were not a part of the task's evaluation rubric. Similarly, the precision metric may falsely consider a claim incorrect since it was not present in the task's ground-truth answer.  Therefore, we follow a hybrid strategy where we manually evaluate all claims generated by all  DR systems under evaluation. If any of the generated claims are correct, we add them to the ground-truth set of claims. After this expansion of the ground-truth, we do an automated evaluation against the expanded ground-truth. Our final benchmark data includes this expanded ground-truth for each question.  

\section{Using \ldr to Evaluate Deep Research Systems}
We present detailed evaluation of state-of-the-art Deep Research models and reasoning and non-reasoning LLMs (as baselines) on our benchmark. We also provide an analysis of the inference-time deep research methodology of these models as gleaned from the output (summarized) reasoning traces.

\subsection{Setup}
We evaluate three proprietary DR models: OpenAI DR \citep{openai-blog}, Perplexity  DR \citep{perplexity-dr}, and Gemini DR (with 2.5 Pro) \citep{gemini-dr}. In addition, we also evaluate three reasoning-enabled baselines: OpenAI’s o4-mini \citep{openai-o4mini}, Perplexity’s Sonar Reasoning~\citep{perplexity-sonar-reasoning}, and Gemini 2.5 Pro \citep{gemini-2.5-family}, and 3 non-reasoning baselines: OpenAI’s GPT-4.1 \citep{openai-gpt41}, Perplexity’s Sonar Pro~\citep{perplexity-sonar-pro}, and Gemini 2.5 Flash~\citep{gemini-2.5-family}. All baselines have \textit{web search} enabled. 

At the time of writing this paper, the proprietary DR models did not have API access. So we evaluate these models in their chat interface by manually feeding in the DR tasks in our benchmark. Typically, DR models in these chat applications respond with clarification questions in the beginning. We let the models use their best judgment with a fixed response to the effect of ``Go Ahead''. 

For the baselines, we rely on the APIs provided by each model's provider. All hyperparameters are kept at their default settings, except for the search context size and reasoning effort/budget parameters. For search context size, we chose ``\texttt{medium}" for OpenAI and Perplexity baselines, with the Gemini API not providing an equivalent parameter. For reasoning effort/budget, we chose ``\texttt{low}" for o4-mini, ``\texttt{medium}" for Sonar Research, and \texttt{128 tokens} for Gemini 2.5 Pro -- each corresponding to $\approx$30 seconds of thinking time by the respective models.  


\subsection{Overall Results}
\label{sec:overall}

\begin{table}[h!]
\small
\renewcommand{\arraystretch}{1.25} 
\centering
\begin{tabular}{
   >{\raggedright\arraybackslash}m{2.0cm}|
   >{\centering\arraybackslash}m{1cm}
   >{\centering\arraybackslash}m{0.8cm}
   >{\centering\arraybackslash}m{0.8cm}|
   >{\centering\arraybackslash}m{1cm}
   >{\centering\arraybackslash}m{0.8cm}
   >{\centering\arraybackslash}m{0.8cm}|
   >{\centering\arraybackslash}m{1cm}
   >{\centering\arraybackslash}m{0.8cm}
   >{\centering\arraybackslash}m{0.8cm}
}
\toprule
\multirow{2}{*}{\textbf{Subset}} &  \multicolumn{3}{c|}{\textbf{OpenAI Deep}}  &  \multicolumn{3}{c|}{\textbf{Perplexity Deep}}  &  \multicolumn{3}{c}{\textbf{Gemini Deep}} \\ 
 &  \multicolumn{3}{c|}{\textbf{Research}}  &  \multicolumn{3}{c|}{\textbf{Research}}  &  \multicolumn{3}{c}{\textbf{Research (2.5 Pro)}} \\ 
& \textbf{Prec} & \textbf{Rec} & \textbf{F1} & \textbf{Prec} & \textbf{Rec} & \textbf{F1} & \textbf{Prec} & \textbf{Rec} & \textbf{F1} \\
\midrule
\scifacts
Materials & 0.313 & 0.316 & \textbf{0.314} & 0.264 & 0.105 & \underline{0.150} & 0.072 & 0.013 & 0.022 \\
\scifacts
Geo & 0.715 & 0.728 & \textbf{0.721} & 0.263 & 0.144 & 0.186 & 0.405 & 0.259 & \underline{0.316} \\
\datasets Identification & 0.667 & 0.667 & \textbf{0.667} & 0.633 & 0.633 & \underline{0.633} & 0.400 & 0.400 & 0.400 \\
\datasets Identification and Extraction & 0.526 & 0.448 & \textbf{0.470} & 0.325 & 0.349 & 0.333 & 0.406 & 0.329 & \underline{0.345} \\
\datasets Peer Retrieval& 0.795 & 0.480 & \textbf{0.585} & 0.494 & 0.248 & 0.311 & 0.551 & 0.249 & \underline{0.338} \\
\rule{0pt}{1em}\patents & 0.694 & 0.463 & \textbf{0.539} & 0.689 & 0.339 & \underline{0.419} & 0.106 & 0.096 & 0.082 \\
\rule{0pt}{1em}\entities & 0.813 & 0.534 & \textbf{0.603} & 0.673 & 0.373 & \underline{0.447} & 0.494 & 0.3 & 0.338 \\
\rule{0pt}{1em}\flights & 0.542 & 0.546 & \textbf{0.540} & 0.389 & 0.347 & \underline{0.362} & 0.249 & 0.276 & 0.261 \\
\midrule
\textbf{Weighted Average} & {0.629} & {0.519} & \textbf{0.550} & {0.462} & {0.286} & \underline{0.331} & {0.309} & {0.215} & {0.236} \\
\textbf{Average} & {0.633} & {0.523} & \textbf{0.555} 
& {0.466} & {0.317} & \underline{0.355} 
& {0.335} & {0.240} & {0.263} \\
\bottomrule
\end{tabular}
\caption{Precision, recall, and F1 scores on \ldr for Deep Research models. Claim agreement score is evaluated using GPT-4o.}
\label{tab:overall}
\renewcommand{\arraystretch}{1}
\end{table}

Table~\ref{tab:overall} reports precision, recall, and F1 scores for all Deep Research models across all task categories. We report both average metrics and weighted average metrics, where the weight is the number of questions in each category. The claim agreement scores are computed using GPT-4o (Section \ref{sec:eval})  and the metrics are computed using Equation \ref{eq:prec-recall}. Across all categories, the OpenAI DR model achieves the best performance, often significantly better than both Perplexity and Gemini DR models, with an overall average F1 or 0.55. While Gemini outperforms Perplexity on specific tasks, such as \scifacts Geo, \datasets Identification and Extraction, \datasets Peer Retreival, Perplexity exhibits stronger average performance compared to Gemini. Across models, the F1 score per category ranges from 0.02 (Gemini DR on \scifacts Materials) to 0.72 (OpenAI DR on \scifacts Geo). These results highlight the difficulty of LiveDRBench, with state-of-the-art models struggling in many of the task categories and leaving a substantial room for improvement. 

The DR models perform worst on \scifacts Materials, \datasets Identification and Extraction, and \flights. This is likely because these tasks require not only identifying specific materials, flights or datasets, but also reasoning across multiple sections of the papers, flight reports to extract the findings required by the queries. The models perform best on \scifacts Geo, likely because this task does not involve any subclaim extraction. Similarly, the average performance on \datasets Identification is higher because of the relative simplicity of the task, since involves just the identification of the paper, its publication venue, and its hosted link--which is all closely related information. For \scifacts materials, we observe that models often correctly extract either the paper title or the material name, but not both, resulting in low overall F1 scores. OpenAI DR achieves an F1 of 0.735 for paper titles and 0.504 for materials, but only 0.314 overall. Similarly, PPL DR reaches 0.348 (title), 0.32 (material), and 0.158 (overall), while Gemini DR performs worst with 0.17, 0.305, and 0.023, respectively. This highlights the difficulty of accurate extraction of both subclaim and main claim.


\xhdr{\textbf{Category-wise Results}} We present category-wise results for the DR models and baselines in Tables \ref{tab:geo}--\ref{tab:entities}. Among the baseline non-DR models, reasoning models struggle when their reasoning is limit to $\approx$30 seconds, underscoring the complexity of LiveDRBench queries -- requiring intensive compute and planning capabilities. In particular, all reasoning models obtain a F1 score of zero for the \scifacts Materials dataset. 
The non-reasoning models perform even worse, due to their limited ability to plan, conduct iterative searches, and backtrack. Interestingly, OpenAI's o4-mini performs comparably or even better than Perplexity DR and Gemini DR in many of the task categories, yet significantly behind OpenAI DR.

To check for errors due to GPT-4o based agreement scores, we also present human-evaluated precision, recall, and F1 scores across all categories in Table~\ref{tab:overall-human} of Appendix ~\ref{app:additional-results}. Overall, we find that the metrics computed using the claim agreement scores determined through manual verification by the authors closely match that of the results in Table~\ref{tab:overall}. This indicates that our benchmark can be used to reliably evaluate any new model using the evaluation prompts in Appendix~\ref{app:evalprompts}.  


\begin{table}[t]
\footnotesize
  \begin{minipage}{.49\linewidth}
    \centering
    \begin{tabular}{lccc}
\toprule
\textbf{Model} & \textbf{Precision} & \textbf{Recall} & \textbf{F1} \\
\midrule
OpenAI DR & 0.715 & 0.728 & \textbf{0.721} \\
Perplexity DR & 0.263 & 0.144 & 0.186 \\
Gemini DR  & 0.405 & 0.259 & \underline{0.316} \\
\midrule
OpenAI o4-mini & 0.138 & 0.104 & \underline{0.114} \\
Sonar Reasoning & 0.000 & 0.000 & 0.000 \\
Gemini 2.5 Pro & 0.273 & 0.200 & \textbf{0.201} \\
\midrule
OpenAI GPT-4.1 & 0.017 & 0.021  & \underline{0.019} \\
Sonar Pro & 0.000 & 0.000  & 0.000 \\
Gemini 2.5 Flash & 0.059 & 0.059 & \textbf{0.059} \\
\bottomrule
\end{tabular}
    \caption{Comparison of DR models on the \scifacts Geo tasks.}
    \label{tab:geo}
  \end{minipage}%
  \quad
  \begin{minipage}{.49\linewidth}
    \centering
    \begin{tabular}{lccc}
\toprule
\textbf{Model} & \textbf{Precision} & \textbf{Recall} & \textbf{F1} \\
\midrule
OpenAI DR & 0.313 & 0.316 & \textbf{0.314} \\
Perplexity DR & 0.264 & \underline{0.105} & 0.150 \\
Gemini DR  & 0.072 & 0.013 & 0.022 \\
\midrule
OpenAI o4-mini & 0.000 & 0.000  & 0.000 \\
Sonar Reasoning & 0.000 & 0.000  & 0.000 \\
Gemini 2.5 Pro & 0.000 & 0.000  & 0.000 \\
\midrule
OpenAI GPT-4.1 & 0.000 & 0.000  & 0.000 \\
Sonar Pro & 0.000 & 0.000  & 0.000 \\
Gemini 2.5 Flash & 0.000 & 0.000 & 0.000 \\
\bottomrule
\end{tabular}
\caption{Comparison of DR models on the \scifacts Materials tasks.}
\label{tab:mpve}
  \end{minipage}%
\end{table}


\begin{table}[t]
\footnotesize
  \begin{minipage}{.49\linewidth}
    \centering
    \begin{tabular}{lccc}

\toprule
\textbf{Model} & \textbf{Precision} & \textbf{Recall} & \textbf{F1} \\
\midrule
OpenAI DR & 0.667  & 0.667  &\textbf{ 0.667}  \\
Perplexity DR & 0.633  & 0.633  & \underline{0.633}  \\
Gemini DR & 0.400  & 0.400  & 0.400  \\
\midrule
OpenAI o4-mini & 0.467  & 0.467  & \textbf{0.467 } \\
Sonar Reasoning & 0.200  & 0.200  & 0.200  \\
Gemini 2.5 Pro & 0.444  & 0.444  & \underline{0.444}  \\
\midrule
OpenAI GPT-4.1 & 0.333   & 0.333   & \underline{0.333}   \\
Sonar Pro & 0.233   & 0.233   & 0.233   \\
Gemini 2.5 Flash & 0.378  & 0.378  & \textbf{0.378}  \\
\bottomrule
\end{tabular}
\caption{Comparison of DR models and baselines on the \datasets Identification tasks.}
\label{tab:datasetsid}
  \end{minipage}%
  \quad
  \begin{minipage}{.49\linewidth}
    \centering
    \begin{tabular}{lccc}
\toprule
\textbf{Model} & \textbf{Precision} & \textbf{Recall} & \textbf{F1} \\
\midrule
OpenAI DR & 0.526 & 0.448 & \textbf{0.470} \\
Perplexity DR & 0.325 & 0.349 & 0.333 \\
Gemini DR & 0.406 & 0.329 & \underline{0.345}  \\
\midrule
OpenAI o4-mini & 0.203 & 0.146 & \textbf{0.168} \\
Sonar Reasoning & 0.015 & 0.003 & 0.005 \\
Gemini 2.5 Pro & 0.186 & 0.130 & \underline{0.142} \\
\midrule
OpenAI GPT-4.1 & 0.126 & 0.078 & \underline{0.088} \\
Sonar Pro & 0.027 & 0.020 & 0.023 \\
Gemini 2.5 Flash & 0.211 & 0.097 & \textbf{0.111} \\
\bottomrule
\end{tabular}
\caption{Comparison of DR models and baselines on the \datasets Identification and Extraction tasks.}
\label{tab:datasetsidex}
  \end{minipage}%
\end{table}

\begin{table}[t]
\footnotesize
  \begin{minipage}{.49\linewidth}
    \centering
\begin{tabular}{lccc}
\toprule
\textbf{Model} & \textbf{Precision} & \textbf{Recall} & \textbf{F1} \\
\toprule
OpenAI DR & 0.795 & 0.480 & \textbf{0.585} \\
Perplexity DR & 0.494 & 0.248 & 0.311  \\
Gemini DR & 0.551 & 0.249 & \underline{0.338} \\
\midrule
OpenAI o4-mini & 0.652 & 0.254 & \textbf{0.345} \\
Sonar Reasoning & 0.083 & 0.009 & 0.017 \\
Gemini 2.5 Pro & 0.334 & 0.140 & \underline{0.196} \\
\midrule
OpenAI GPT-4.1 & 0.465 & 0.196 & \textbf{0.276} \\
Sonar Pro & 0.231 & 0.083 & 0.122 \\
Gemini 2.5 Flash & 0.229 & 0.163 & \underline{0.190} \\
\bottomrule
\end{tabular}
\caption{Comparison of DR models and baselines on the \datasets Peer Retrieval tasks.}
\label{tab:datasetspr}
  \end{minipage}%
  \quad
  \begin{minipage}{.49\linewidth}
    \centering
\begin{tabular}{lccc}
\toprule
\textbf{Model} & \textbf{Precision} & \textbf{Recall} & \textbf{F1} \\
\midrule
OpenAI DR & 0.694 & 0.463 & \textbf{0.539} \\
Perplexity DR & 0.689 & 0.339 & \underline{0.419} \\
Gemini DR & 0.106 & 0.096 & 0.082 \\
\midrule
OpenAI o4-mini & 0.529 & 0.179 & \underline{0.254} \\
Sonar Reasoning & 0.230 & 0.149 & 0.166 \\
Gemini 2.5 Pro & 0.309 & 0.322 & \textbf{0.307} \\
\midrule
OpenAI GPT-4.1 & 0.362 & 0.297 & \textbf{0.311} \\
Sonar Pro & 0.180 & 0.100 & 0.113 \\
Gemini 2.5 Flash & 0.272 & 0.228 & \underline{0.227} \\
\bottomrule
\end{tabular}
\caption{Comparison of DR models and baselines on the \patents tasks.}
\label{tab:patents}
  \end{minipage}%
\end{table}

\begin{table}[h]
\footnotesize
  \begin{minipage}{.49\linewidth}
    \centering
    \begin{tabular}{lccc}
\toprule
\textbf{Model} & \textbf{Precision} & \textbf{Recall} & \textbf{F1} \\
\toprule
OpenAI DR & 0.542 & 0.546 & \textbf{0.540} \\
Perplexity DR & 0.389 & 0.347 & \underline{0.362} \\
Gemini DR & 0.249 & 0.276 & 0.261 \\
\midrule
OpenAI o4-mini & 0.340 & 0.282 & \textbf{0.304} \\
Sonar Reasoning & 0.237 & 0.168 & 0.183 \\
Gemini 2.5 Pro & 0.228 & 0.210 & \underline{0.215} \\
\midrule
OpenAI GPT-4.1 & 0.200 & 0.138 & \underline{0.160} \\
Sonar Pro & 0.216 & 0.187 & \textbf{0.194} \\
Gemini 2.5 Flash & 0.182 & 0.105 & 0.117 \\
\bottomrule
\end{tabular}
\caption{Comparison of DR models and baselines on the \flights tasks.}
\label{tab:flights}
  \end{minipage}%
  \quad
  \begin{minipage}{.49\linewidth}
    \centering
\begin{tabular}{lccc}
\toprule
\textbf{Model} & \textbf{Precision} & \textbf{Recall} & \textbf{F1} \\
\toprule
OpenAI DR &0.813 & 0.534 & \textbf{0.603}

 \\
Perplexity DR &0.673 & 0.373 & \underline{0.447}

 \\
Gemini DR &0.494 & 0.3 &	0.338
\\
\midrule
OpenAI o4-mini &0.265 & 0.082 & \underline{0.115}  \\ 
Sonar Reasoning & 0.199& 0.04 & 0.064\\ 
Gemini 2.5 Pro &  0.254 &  0.120 &    \textbf{0.151} \\ 
\midrule
OpenAI GPT-4.1 &0.217 & 0.0725 &\textbf{ 0.074} \\ 
Sonar Pro &0.075 &0.035 & 0.042 \\ 
Gemini 2.5 Flash & 0.153 &  0.044 &   \underline{0.064} \\ 
\bottomrule
\end{tabular}
\caption{Comparison of DR models on the \entities tasks.}
\label{tab:entities}
    
  \end{minipage}%
\end{table}

\subsection{Analysis of Deep Research Models}

We now analyze the reasoning traces produced by deep research models. We choose the \datasets category for this analysis.  For each query, we extract the corresponding traces from the deep research models and quantify three key behaviors: \textbf{(1)} the number of \textbf{sources} referenced, \textbf{(2)} the number of \textbf{backtracking} events, and \textbf{(3)} the number of \textbf{branches}. We define backtracking as any instance where the model revisits, revises, or changes its line of reasoning. Similarly, we define branches as the number of distinct research directions, hypotheses, or subgoals pursued within a trace. Finally, the number of sources is defined as the total number of websites visited by the DR model in order to address a given query. 

\begin{figure}[t]
    \centering
    \includegraphics[width=0.98\linewidth]{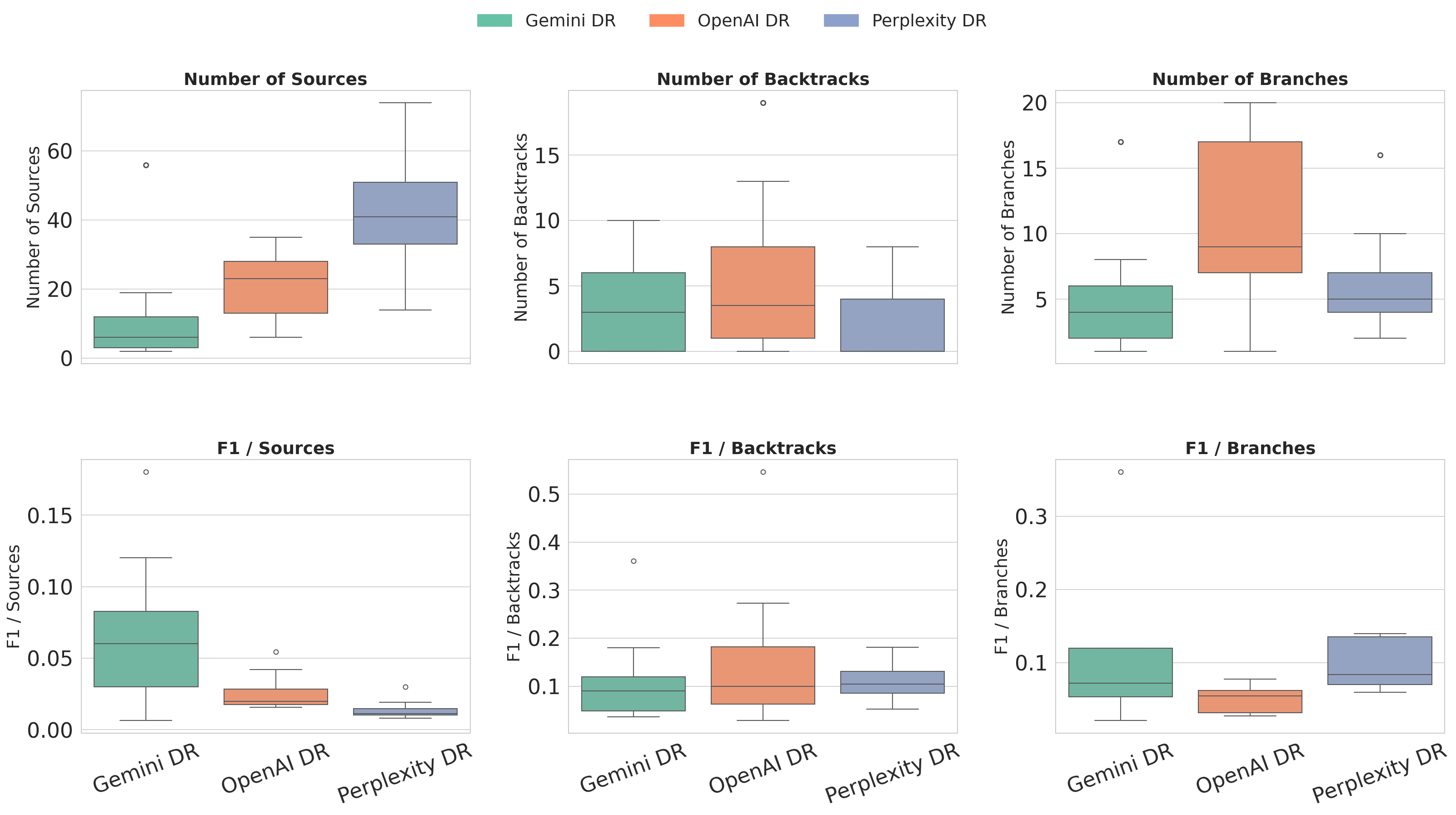}
    \label{fig:fig3}

    \caption{(Top) Box Plots illustrating the number of \textbf{sources} referred, \textbf{branches} considered, and \textbf{backtracking} events occurred against F1 score on the \datasets subset using three DR models: Gemini DR, OpenAI DR (OAI DR), Perplexity DR. (Bottom) F1 Efficiency per event.}
    \label{fig:trace-anal}
    
\end{figure}

We extract the number of backtracking operations and branches by providing the entire trace to GPT-4o~\citep{gpt-4o} and explicitly prompting it. We provide detailed prompts to extract this information in the Appendix~\ref{sec:app-trace}. The total number of sources are extracted by regex pattern matching since the DR model traces typically contain this information. In Figure~\ref{fig:fig3}, we plot the mean and standard deviation of these characteristics on 19 queries from the \datasets category \footnote{We could not extract the trace from the chat interface for one query.}, alongside the average F1 score.


\xhdr{Overall Results} 
OpenAI DR (F1=0.55) outperforms both Perplexity DR (F1=0.41) and Gemini DR (F1=0.35) in terms of average F1 score on this subset. We note that OpenAI DR generates the largest number of branches and backtracks the most on average, which likely contributes to better F1 score. We also note that gathering large number of sources does not necessarily correlate with higher F1 as evidenced by the results on Perplexity DR.


\xhdr{LiveDRBench is Challenging}
Figure~\ref{fig:trace-anal} (Top) shows the average number of referenced sources, branches, and backtracking events in the reasoning traces for each model. It highlights the complexity of the \datasets Identification and Extraction task, where all three models---OAI DR, Gemini DR, and Perplexity DR---engage in complex reasoning workflows involving a large number of sources, as well as frequent backtracking and branching. For example, Perplexity DR and OAI DR utilize an average of over 40 and 20 sources respectively, yet achieve average F1 scores of only 0.41 and 0.55. This suggests that even with extensive source gathering, the tasks remain difficult. On average, all DR models exhibit broader reasoning, i.e., they tend to branch more than they backtrack. 


\xhdr{Comparing Efficiency}
To evaluate efficiency, we ask: \textit{Which DR model delivers the best performance per unit of reasoning/search effort}? We formalize this by computing F1 score normalized by trace characteristics such as the number of sources, branches, and backtracking operations. As shown in the box plots of Figure~\ref{fig:trace-anal} (bottom), Gemini DR demonstrates the highest efficiency in terms of F1 score per source gathered, but yields the lowest overall F1 score. In contrast, OpenAI DR engages in more extensive reasoning via backtracking and branching, which contributes to greater efficiency than Perplexity DR in terms of F1 per source. These patterns suggest that Gemini DR tends to terminate early, often before sufficient exploration has occurred resulting in fewer retrieved sources and limited reasoning. Perplexity DR, on the other hand, retrieves many sources but performs relatively little backtracking, indicating an exploratory yet shallow approach. Finally, while OpenAI DR achieves the highest average F1 score, its overall efficiency is lower due to frequent branching and backtracking, which incur higher costs.


\section{Discussion and Conclusion}
We presented a characterization of the deep research problem that is amenable to objective evaluation. It separates the search and reasoning subtasks from long form generation and we argued why the former is the defining element of DR. That is, deep research can be defined as the discovery of claims and their supporting evidence, that ultimately form the final long-form answer. Based on this definition, we presented a challenging benchmark of 100 problems and evaluated existing DR systems. 

Our evaluation points to a few directions for improving DR systems. The first is that DR systems do not perform well even when the algorithm is straightforward (but the search may be laborious), e.g., in the \entities benchmark category. This suggests that DR models may be trained on such tasks and finetuned to output an algorithm and follow it. Another option may be to interleave programmatic and model control for such problems. For other problems such as \scifacts, DR models would often yield the correct answer but wrong grounding or vice-versa. Some amount of error is inevitable when processing large corpora; additional tools that can operate on the reasoning trace and update the  model-generated report can be useful. Finally, our analysis shows the extent of branching and backtracking operations executed by the model. Understanding the tradeoffs of these operations will be important for improving the inference-time efficiency of deep research models.

\bibliographystyle{plainnat}
\bibliography{dr}

\appendix
\section{Example tasks from \ldr}
\label{app:examples}
In this section, we provide examples from each category (Table~\ref{tab:tasks-eg}) in the \ldr.

\begin{tcolorbox}[title=\scifacts Materials Prompt]
Find the material(s) that satisfy every one of the listed measured properties.\\

Instructions  \\
1. In your response, provide a list of JSON object(s) in the following format:  \\
\begin{lstlisting}[language=json]
[  
    {  
      "material": "<material name>",  
      "inference_basis": "<brief explanation of how the listed properties match this material>",  
      "paper_title": "<title of the paper>",  
      "property_source_table": "<source table from which material name, property and property descriptor were identified>",  
      "property_source_passage": "<source passage from which material name, property and property descriptor were identified>"  
    }  
  ]
\end{lstlisting}
2. Your inference\_basis should reference the key properties (e.g., band gap, absorption coefficient, crystal structure) that led you to that material.  \\
3. You must provide a valid json file even if you don't have the answer.\\

Measured properties:  \\
- Band gap Eg (room temperature): 3.35 eV  \\
- binding energy: 60 meV  \\
- Lattice constant a (X-ray diffraction XRD): 0.325 nm \\  
- Lattice constant c (X-ray diffraction XRD): 0.521 nm  \\
- Lattice fringe (High-resolution transmission electron microscopy HRTEM): 0.52 nm
\end{tcolorbox}

\begin{tcolorbox}[title=\scifacts Geo Prompt]
Find the paper(s) that use all of the datasets mentioned below: \\ 
Google Earth Engine, Shuttle Radar Topography Mission data, Ground Truth Data, Moderate Resolution Imaging Spectroradiometer (MODIS) Data
\\
Instructions:  \\
1. You must respond with a list JSON object(s) in this format:  
\begin{lstlisting}[language=json]
{
    "paper_title": "<full paper title>"
}
\end{lstlisting}
2. You must provide a valid json file with empty fields even if you don't have the answer.
\end{tcolorbox}

\begin{tcolorbox}[title=\datasets Identification Prompt]
I'm looking for diffusion based approaches that enable precise control over illumination in images, particularly those that move beyond simple prompt engineering. I’m especially interested in a method that uses a diverse image collection, and incorporates physically grounded constraints—such as consistency under blended lighting conditions—during training, to allow models to respect the structural coherence of light transport. I’m especially curious about methods where the consistency is strong enough to infer structural cues like surface normals purely from illumination variation across synthetic views, without being specifically trained for this property. Find a paper, ideally a published one, that does this.
\\
Once you've found a paper that meets my needs, please provide the information in the following format:\\

\begin{lstlisting}[language=json]
{
    'title': <title>,
    'venue': <venue>,
    'year': <year>,
}
\end{lstlisting}
\end{tcolorbox}

\begin{tcolorbox}[title=\datasets Identification and Extraction]
I'm looking for a GPT-4 generated corpus of decisions rooted in quotidian life - think commuting, family squabbles or career decisions - each tagged against broad socio-psychological dimensions. I'm specifically interested in scenarios where the resolution hinges not on objective correctness, but instead on personal values. How does this corpus reveal GPT-4’s implicit generation bias across the various value dimensions in each of the socio-psychological frameworks explored in the corpus? \\

Provide the an overview of GPT-4's generation bias in the following json format: \\

\begin{lstlisting}[language=json]
[
    {
        "framework": <framework_name>,
        "most_biased_dimension": <most_biased_dimension_name>
    },
    ...
]
\end{lstlisting}
\end{tcolorbox}

\begin{tcolorbox}[title=\datasets Peer Retrieval]
I want to compare existing 3D urban segmentation datasets based on real-world scenes that go beyond static 2D maps or flat segmentation overlays, and instead represent urban environments as richly annotated point clouds. How do they compare in terms of data acquisition methods—such as mobile laser scanning, terrestrial laser scanning, aerial laser scanning, and photogrammetry? \\

For the datasets you find, provide the following information in json format: \\

\begin{lstlisting}[language=json]
[
    {
        "name": <name of the resource>,
        "data_aquisition_method": <data acquisition method>,
        "area": <area of the dataset>,
        "scenes": <number of scenes>,
        "points_million": <number of points>,
    },
    ...
]
\end{lstlisting}
\end{tcolorbox}

\begin{tcolorbox}[title=\entities Prompt]
List all movies released between **2005 and 2011** (inclusive) that have **won an Oscar in any category**, and are **adaptations of works authored (or co-authored) by women**.
\\
Please provide the answer in **JSON format**, where each entry includes the following three fields:\\

- ``Movie Name" – the title of the film  \\
- ``Original Work" – the name of the source material the movie is based on  \\
- ``Author Name" – the name(s) of the female author(s) of the original work
\end{tcolorbox}

\begin{tcolorbox}[title=\flights Prompt]
In which high-profile flight investigation was military radar data later considered unreliable? Identify the types of radar data that was recorded, what anomalies were found, and why those anomalies raised doubts. Focus on reconstructing a timeline of anomalous readings and critically examine the underlying causes.  \\
For each data type and anomaly, follow this format:\\

\begin{lstlisting}[language=json]
[
    {
        "timestamp_utc": "<timestamp>",
        "parameter": "<parameter_name>",
        "recorded_value": "<value>",
        "expected_range_or_behavior": "<expected_range_or_behavior>",
        "anomaly_description": "<description_of_anomaly>"
    },
    ...
]
\end{lstlisting}
\end{tcolorbox}

\begin{tcolorbox}[title=\patents Prompt]
I have the following ideas for a research paper. Can you help identify if this has already been done or implied in full or in parts in other papers? \\
Give your answer as a **JSON with three fields: Paper title, link, and connection a field** that quotes exact sentences from the paper and parts of my ideas below to make the case.\\

Here is the format:\\

\begin{lstlisting}[language=json]
[
    {
        "title": <paper title>,
        "link": <link to the paper>,
        "connection": <connection field>
    }
]
\end{lstlisting}
The objective of our paper is to understand optimized prompts (e.g. using black-box or white-box prompt optimization techniques) better in terms of how they differ from human-written forms, and how they help LMs do better predictions, sometimes therefore also enabling jailbreaking, etc. We formulate adversarial attack as a form of controllable text generation. Our method injects attack information during the decoding steps, enabling successful attacks.\\

The key idea is to automatically generate a “suffix” (a sequence of tokens) to concatenate to the user input (user prompt, task description, etc.).
The tokens will be generated one by one left-to-right by maximising a mixed search objective, that uses gradient in the vocabulary and also account for readability of the prompt. Readability in the objective is important especially because a common complaint against optimized prompts is that they often contains illegible text, punctuations, etc.\\
Once we have such an optimized readable prompt, we argue that it consists of influential, specific, or rare tokens that help better elicit the user behavior, or in some cases even malicious behavior in case of a jailbreaking attempt. Further, we show that the optimized prompts have distinct internal embeddings (using sparse probing techniques), not just lexical behaviors. Notably, we show results on two new attack scenarios (inducing system prompt leakage and addressing over-censored tasks in LLMs) that have not been explored before.
\end{tcolorbox}
\section{Evaluation Prompts}
\label{app:evalprompts}
Here, we provide the exact prompts used for evaluating the generated outputs. Across all our experiments, we leverage GPT-4o~\cite{gpt-4o} as the judge model and compare the generated answer with the ground truth.To ensure the ground truth is comprehensive, we conduct additional human checks and augment it with valid answers that were initially not present. Our benchmark is built on this enriched ground truth set.

For \scifacts Materials task we evaluate each key (paper title and material name) present in the generated list of dictionaries against the prompt given below and report the combined metric according to Eq.~\ref{eq:prec-recall}. We use the same evaluation prompt to evaluate the paper title in \scifacts Geo task.

\begin{tcolorbox}[title=Prompt for evaluating each key in SciFacts]
You are an evaluator. Given a Golden Answer and a Predicted Answer, check whether the Predicted Answer is correct. The prediction is correct if it fully aligns with and contains all the key information present in the Golden Answer. Respond with True if the prediction is correct and False otherwise.\\
Golden Answer: \{\texttt{golden-answer}\}\\
Predicted Answer: \{\texttt{predicted-answer}\}
\end{tcolorbox}

For the \datasets and \flights queries in which both the prediction and ground truth consist of lists of dictionaries, we begin by identifying the corresponding ground-truth dictionary for each predicted dictionary. This is done using a set of primary keys to be used for matching, using the following prompt:

\begin{tcolorbox}[title=\datasets and \flights prompt to match a predicted dictionary against a list of ground truth dictionaries using a set of primary keys]
You are given a list of dictionaries (the ground truth) and a single predicted dictionary (the model output). Your task is to determine which dictionary in the ground truth list corresponds with the predicted dictionary, using a set of primary keys to identify the best match. \\

- You will be a given a list of primary keys that can be used to compare the dictionaries. If a primary key is not present in two dictionaries, make sure to use the other primary keys to determine the match. \\
- Do not consider ANY other fields in the dictionaries except for the primary keys.
- If the value is a string, consider them equivalent if they are similar in meaning, or if one is a subset of the other. \\
- Match based on semantic similarity or if one is a subset of the other, not exact string equality. \\
    - Acronyms, shortened names, or partial names are equivalent to their expanded forms. \\
    - "BERT" and "BERT: Pre-training of Deep Bidirectional Transformers for Language Understanding" $\xrightarrow[]{}$ equivalent \\
    - "ICLR" and "International Conference on Learning Representations" $\xrightarrow[]{}$ equivalent \\
    - A prediction may include extra info (e.g., a year or qualifier). If the ground truth is a subset and meaning is preserved, treat them as equivalent. If the ground truth includes a qualifier, the prediction's qualifier must match if present. \\
- You may also be provided with a set of extra evaluation notes that provide more specific evaluation criteria. These extra rules may override the general rules above. \\
- If after all this no match exists, return an empty dictionary. \\

Now compare the following: \\

Ground Truth List: \\
\{\texttt{list-gt}\} \\

Model Output: \\
\{\texttt{dict-pred}\} \\

Primary Keys: \\
\{\texttt{primary-keys}\} \\ 

Extra Evaluation Notes: \\
\{\texttt{note}\} \\

Your output should be the dictionary from the ground truth list that best matches the model output, or an empty dictionary if no match is found. \\

IMPORTANT: \\
- Do NOT explain your reasoning. \\
- Do NOT include any extra text. \\
- Only output the final dictionary as a JSON object. Do not wrap it in markdown or say anything else.
\end{tcolorbox}

With a corresponding ground-truth dictionary obtained for each predicted dictionary in the \datasets and \flights queries, we evaluate the entire predicted dictionary against the ground truth dictionary, rather than key by key. This is because related sub-claims provide important context, and evaluating keys individually can lead to under-scoring by the LLM judge. The exact prompt used for evaluation is provided below:

\begin{tcolorbox}[title=\datasets and \flights prompt for evaluating  a predicted dictionary against a ground truth dictionary]
You are evaluating whether two dictionaries refer to the same information. Your task is to determine whether each corresponding value is equivalent. \\

For each key-value pair in the ground truth dictionary:  \\
- If the key is absent in the model output dictionary, it is not equivalent. \\
- If the value is a number (integer or float), they are equivalent within a 1\% margin of error. \\
- If the value is a string, consider them equivalent if they are similar in meaning, or if one is a subset of the other. \\
    - Acronyms or shortened names are equivalent to their expanded forms. \\
    - "BERT" and "BERT: Pre-training of Deep Bidirectional Transformers for Language Understanding" $\xrightarrow{}$ equivalent \\
    - "ICLR" and "International Conference on Learning Representations" $\xrightarrow{}$ equivalent \\
    - "CNN" and "ResNet" $\xrightarrow{}$ equivalent (both are convolutional neural networks) \\
    - "Adam" and "SGD" $\xrightarrow{}$ not equivalent (different optimizers) \\
    - "0.8" and "0.80" $\xrightarrow{}$ equivalent numerically \\
- If the value is a URL, they must be exactly the same, or a shortened version of the same URL. \\
- If the value is a list, check if every element in the ground truth list is present in the model output list, regardless of order. Follow the same rules for strings as above. \\
- If the value is a dictionary, check if every key-value pair in the ground truth dictionary is present in the model output dictionary, regardless of order. Again, follow the same rules for strings as above. \\
- Some values may depend on information from other keys in the dictionary. Use such context to determine whether two values are equivalent, even if they are not identical on their own. \\
- Ignore trivial formatting differences such as casing, whitespace, or common abbreviations if the meaning is preserved. \\

Now compare the following dictionaries: \\

Ground Truth: \\
\{\texttt{dict-gt}\} \\

Model Output: \\
\{\texttt{dict-pred}\} \\

Your output should be a JSON object with the same keys as the input dictionaries. For each key, the value should be: \\
- 3: The predicted value closely matches the ground-truth, preserving almost all important information with minimal or no loss of meaning. \\
- 2: The predicted value captures the main idea of the ground-truth but omits key details or has minor inaccuracies. \\
- 1: The predicted value shows some overlap with the ground-truth but misses most of the essential meaning. \\
- 0: The predicted value is largely incorrect, unrelated, or does not correspond meaningfully to the ground-truth value. \\

IMPORTANT: \\
- Do NOT explain your reasoning. \\
- Do NOT include any extra text. \\
- Only output the final JSON object. Do not wrap it in markdown or say anything else. \\
\end{tcolorbox}

Finally, we evaluate \patents and \entities using the prompt below. Here, the LLM judge is required to check if the predicted string is present in the list of ground truths (paper titles and entities like names of people or movies). We ensure the minor differences in formatting are ignored. 

\begin{tcolorbox}[title=\patents and \entities prompt to match a predicted string against a list of ground truth strings]
You are evaluating whether a predicted string is present in a ground-truth list of strings. \\

- Allow for minor differences in formatting, such as casing, whitespace, or common abbreviations. \\
- If found, return the matching string from the ground truth list in JSON format: \\    \texttt{\{ "match": "matched-string" \}} \\
- If not, return an empty dictionary. \\

Now compare the following: \\

Ground Truth List: \\
\{\texttt{gt-list}\} \\

Predicted String: \\
\{\texttt{pred-str}\} \\

IMPORTANT: \\
- Do NOT explain your reasoning. \\
- Do NOT include any extra text. \\
- Only output the JSON object. Do not wrap it in markdown or say anything else.
\end{tcolorbox}

\section{Trace Analysis Prompts}
\label{sec:app-trace}
\begin{tcolorbox}[title=Prompt to count the number of branches in a given DR trace]
You are an expert in analyzing research investigation traces. Your task is to read the provided trace and identify how many distinct research *branches* or lines of inquiry the author explores. A new branch is a distinct hypothesis, search direction, or subgoal. Count how many times the researcher changes focus, searches for different types of information, or explores parallel possibilities.\\

Output ONLY the number of branches as an integer.
\end{tcolorbox}

\begin{tcolorbox}[title=Prompt to count the number of backtracking events in a given DR trace]
You are an expert in analyzing research investigation traces. Your task is to read the provided trace and count how many times the researcher *backtracks* — that is, revisits or reconsiders previous ideas, changes direction after realizing an approach is suboptimal, or discards a line of inquiry.\\

Look for transitions like:\\
- 'I first thought... but then...'\\
- 'Initially considered... later found...'\\
- 'Switched from... to...'\\

There may or may not be any transition phrases, just count the number of ideas or directions changed.\\

Count only actual *changes in direction or reconsiderations*. Output ONLY the number of backtracking steps as an integer.
\end{tcolorbox}

\section{Additional Results}
\label{app:additional-results}

\begin{table}[h!]
\renewcommand{\arraystretch}{1.5} 
\centering
\begin{tabular}{
   >{\raggedright\arraybackslash}m{2.25cm}|
   >{\centering\arraybackslash}m{1.1cm}
   >{\centering\arraybackslash}m{0.9cm}
   >{\centering\arraybackslash}m{0.9cm}|
   >{\centering\arraybackslash}m{1.1cm}
   >{\centering\arraybackslash}m{0.9cm}
   >{\centering\arraybackslash}m{0.9cm}|
   >{\centering\arraybackslash}m{1.1cm}
   >{\centering\arraybackslash}m{0.9cm}
   >{\centering\arraybackslash}m{0.9cm}
}
\toprule
\multirow{2}{*}{\textbf{Subset}} &  \multicolumn{3}{c|}{\textbf{OpenAI Deep}}  &  \multicolumn{3}{c|}{\textbf{Perplexity Pro}}  &  \multicolumn{3}{c}{\textbf{Gemini Deep}} \\ 
 &  \multicolumn{3}{c|}{\textbf{Research}}  &  \multicolumn{3}{c|}{\textbf{Research}}  &  \multicolumn{3}{c}{\textbf{Research (2.5 Pro)}} \\ 
& \textbf{Prec} & \textbf{Rec} & \textbf{F1} & \textbf{Prec} & \textbf{Rec} & \textbf{F1} & \textbf{Prec} & \textbf{Rec} & \textbf{F1} \\
\hline
\scifacts

Geo & 0.715 & 0.728 & 0.721 & 0.263 & 0.144 & 0.186 & 0.405 & 0.259 & 0.316 \\
\scifacts

Materials & 0.313 & 0.316 & 0.314 & 0.264 & 0.105 & 0.150 & 0.072 & 0.013 & 0.022 \\
\datasets Identification & 0.667 & 0.667 & 0.667 & 0.625 & 0.625 & 0.625 & 0.375 & 0.375 & 0.375 \\
\datasets Identification and Extraction & 0.559 & 0.428 & 0.462 & 0.411 & 0.351 & 0.358 & 0.390 & 0.284 & 0.319 \\
\datasets Peer Retrieval & 0.849 & 0.498 & 0.609 & 0.535 & 0.248 & 0.322 & 0.558 & 0.241 & 0.334 \\
\rule{0pt}{1em}\patents & 0.714 & 0.473 & 0.552 & 0.701 & 0.349 & 0.430 & 0.174 & 0.143 & 0.133 \\
\rule{0pt}{1em}\entities & 0.813 & 0.534 & 0.603 & 0.673 & 0.373 & 0.447 & 0.494 & 0.3 & 0.338 \\
\rule{0pt}{1em}\flights & 0.525 & 0.530 & 0.523 & 0.385 & 0.358 & 0.369 & 0.243 & 0.264 & 0.252 \\
\hline
\textbf{Average} & 0.644 & 0.522 & 0.556 & 0.482 & 0.319 & 0.361 & 0.339 & 0.235 & 0.261 \\
\bottomrule
\end{tabular}
\caption{Precision, recall, and F1 scores on \ldr for Deep Research models. Claim agreement score is evaluated by the authors.}
\label{tab:overall-human}
\renewcommand{\arraystretch}{1}
\end{table}


\end{document}